\documentclass[11pt]{article}

\usepackage[final]{acl}

\usepackage{times}
\usepackage{latexsym}
\usepackage{amsmath}

\usepackage[T1]{fontenc}

\usepackage[utf8]{inputenc}

\usepackage{microtype}

\usepackage{inconsolata}

\usepackage{graphicx}
\usepackage{multicol}
\usepackage{multirow}
\usepackage{algorithm}
\usepackage{algpseudocode}
\usepackage{booktabs} 
\usepackage[skins,breakable]{tcolorbox}
\usepackage{wrapfig}
\definecolor{light-purple}{RGB}{151,156,171}
\definecolor{blue-color}{RGB}{40,166,189}
\definecolor{pink-color}{RGB}{237,46,104} 
\definecolor{dark-grey-color}{RGB}{79,91,102}

\newtcolorbox[list inside=prompt,auto counter,number within=section]{prompt}[1][]{
    colbacktitle=black!80,
    colframe=black!80,
    coltitle=white,
    fonttitle=\sffamily\bfseries,
    fontupper=\ttfamily,
    boxsep=5pt,
    left=0pt,
    right=0pt,
    top=0pt,
    bottom=0pt,
    boxrule=1pt,
    enhanced, 
    breakable,
    skin first=enhanced,
    skin middle=enhanced,
    skin last=enhanced,
    #1,
}

\usepackage[textsize=scriptsize]{todonotes}

\title{Learning Composable Chains-of-Thought}

\author{Fangcong Yin$^{\clubsuit}$, Zeyu Leo Liu$^{\spadesuit\heartsuit}$, \textbf{Liu Leqi}$^{\spadesuit}$, \textbf{Xi Ye}$^{\diamondsuit}$,
\textbf{Greg Durrett}$^{\clubsuit}$ \\
$^{\clubsuit}$New York University, 
$^{\spadesuit}$The University of Texas at Austin \\
$^{\heartsuit}$Salesforce AI Research, 
$^{\diamondsuit}$Princeton University  \\
\url{fy666@nyu.edu}
}

\begin{document}
\maketitle
\begin{abstract}
A common approach for teaching large language models (LLMs) to reason is to train on chains-of-thought (CoTs) of in-distribution reasoning problems. However, such annotated data is costly to obtain for every problem of interest.
We want reasoning models to generalize beyond their training distribution, and ideally to generalize compositionally: they should combine atomic reasoning skills to solve harder unseen tasks. In this paper, we introduce a method to enable generalization to a target compositional task that has no labeled CoT data. We minimally modify CoT formats of constituent atomic tasks to be \textbf{composable}: we add prefixes to CoTs during training time, thereby encouraging the trained model to generate sequences of CoTs during inference. We train individual models on the atomic tasks with composable CoT data and combine them with multitask learning or model merging to address the target compositional task zero-shot. This model can be further trained on a small amount of compositional data using rejection sampling fine-tuning (RFT). Training LLMs on Composable CoT outperforms multitask learning and continued fine-tuning baselines on three domains of controlled compositional tasks and also yields improvement in math skill composition setting derived from the MATH dataset.\footnote{Code and data are available at: \url{https://github.com/fc2869/composable_cot}.}
\end{abstract}

\begin{figure*}
\begin{center}
  \centering
  
  \includegraphics[width=.98\textwidth,trim=10mm 95mm 10mm 35mm,clip]{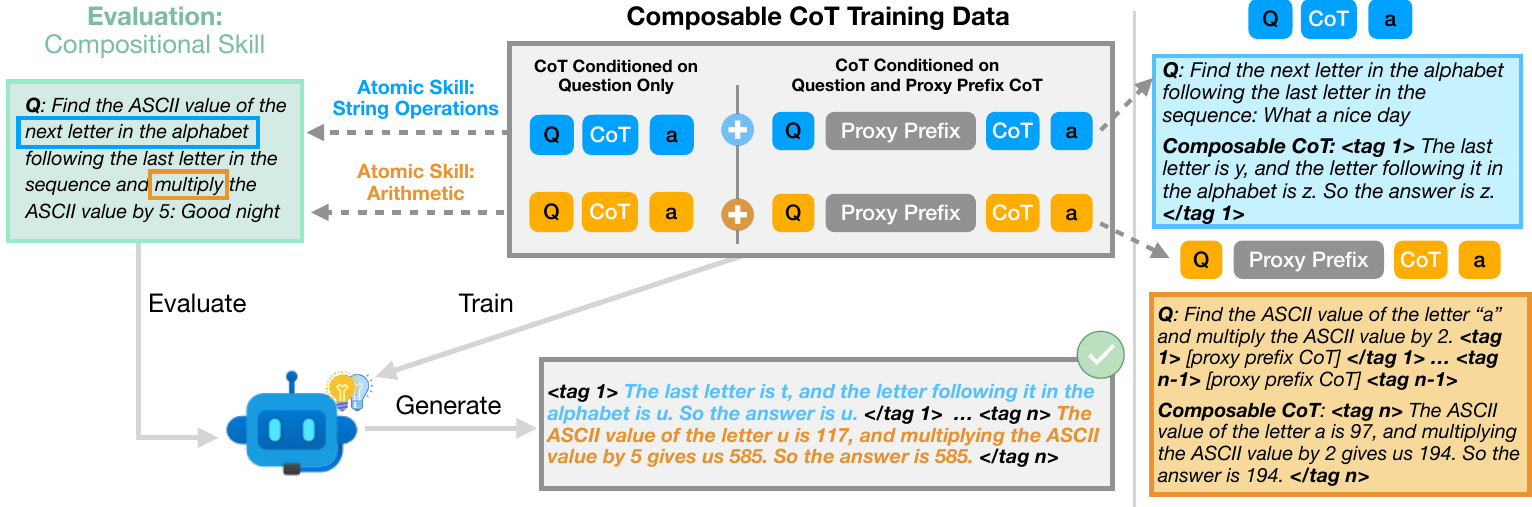}
  
\end{center}
\footnotesize

\caption{\label{fig:composable_cot}  A compositional task involving two atomic skills. We use a data augmentation scheme called \textbf{Composable CoT}: we add ``proxy prefix CoTs,'' such as random filler strings, to the prompt to simulate compositional distributions where a CoT is conditionally generated from the question and other CoTs. At inference time, our model can generate a CoT combining two atomic skills, despite only training on each in isolation. 
}

\end{figure*}
\section{Introduction}

Large language models (LLMs) are successful by virtue of the massive amounts of data they are trained on, which makes a wide range of complex problems in-distribution. However, these models still fail at challenging reasoning tasks, particularly when these reasoning tasks are out-of-distribution. Ideally, we want models that can \emph{generalize} to new settings, especially by applying basic ``skills'' learned during training in novel combinations to solve problems at inference time. How to empower LLMs with this capability, also called compositional generalization \citep{Piantadosi216composition,Werchan20158MonthOldIS,conklin-etal-2021-meta,dziri2023faith}, remains an open question. For instance, large reasoning models \citep{qwq32b,guha2025openthoughtsdatarecipesreasoning},
built on pre-trained LLMs, are typically trained on a large amount of data annotated with chain-of-thought (CoT) traces,
but still fall short at generalizing to harder problems than those they were trained on \citep{sun2024easytohard,hase-etal-2024-unreasonable,abreu2025a,illusion-of-thinking,sun2025omega} even when they leverage CoT reasoning. 

We explore the setting of compositional reasoning where pre-trained LLMs are fine-tuned on CoT data of simple reasoning tasks (atomic tasks) and then evaluated on the \emph{unseen} combinations of them (compositional tasks) with no or limited compositional supervision. We find that models trained with atomic CoT data demonstrate limited generalization to compositional settings. As illustrated in Figure~\ref{fig:composable_cot}, we propose a simple modification of the CoT format of the atomic task training data, which we call \textbf{Composable CoT}: we add ``proxy prefixes,'' which are random filler strings, to the prompt. Then we train models to reason about atomic tasks in the context of the proxy prefixes. Crucially, this training encourages models to chain sequences of CoTs together, which directly enables compositional generalization on a certain class of problems: if a model can chain together a CoT for skill A with a CoT for skill B, it can solve a compositional task requiring skills A and B. 

We first experiment with \emph{zero-shot} combination of Composable CoT models with two different approaches: first, merging models trained on individual atomic CoT tasks, and second, multitask learning across our atomic CoT datasets. Such combined models achieve zero-shot compositional generalization without seeing compositional data during training.

We then demonstrate that our zero-shot models can be improved further by rejection sampling fine-tuning (RFT) on a limited amount of compositional supervision. On controlled compositional tasks involving core reasoning capabilities of LLMs, including natural language skill composition, string manipulation, and arithmetic, our approach outperforms multi-task learning and continued fine-tuning baselines within a given budget of training data. Combining atomic models trained with Composable CoT is better than combining Standard CoT models across different compositional tasks. 

Composable CoT works to combine both pairs and triples of skills in our experiments. 
Moreover, Composable CoT can improve math reasoning using skill composition: training models on Composable CoT data of skills for particular mathematical subjects leads to an average improvement of 4.3\% over the base model on the MATH \citep{hendrycks2021measuring} dataset.

The main contributions of this work include: (1) A novel data augmentation scheme for training CoT models on basic reasoning skills to enable future composition of them for more complicated reasoning tasks. 
(2) A method for improving compositional reasoning with LLMs by first training models on atomic CoTs with such augmentation and then performing rejection sampling finetuning.

\section{Preliminaries}

\textbf{LLM reasoning with chain-of-thought.} 
Given a prompt $\mathbf{q}$ that states a reasoning problem with ground truth answer $a$, an LLM $M$ reasons with chain-of-thought by generating a response that includes a chain-of-thought trace $\mathbf{t}$ followed by a predicted answer $\tilde{a}$. Recent works show that supervised fine-tuning LLMs on CoT traces leads to strong reasoning models \citep{muennighoff2025s1simpletesttimescaling,guha2025openthoughtsdatarecipesreasoning}. We define a dataset for a task $\mathcal{T}$ with CoT traces of size $N$ as a set of (prompt, CoT, answer) triples: $D^{\mathrm{CoT}}_{\mathcal{T}} = \{(\mathbf{q},\mathbf{t},a)\}$.

\textbf{Atomic and compositional tasks.}
Consider a set of tasks that represent basic reasoning skills, which we call \textbf{atomic} tasks. We define \textbf{compositional} tasks  $\mathcal{T}_{A}$ as those tasks that can expressed as a composition of $n$ atomic tasks: $\mathcal{T}_{A} = g(A)$  where $A = \{{\mathcal{T}_{i},...,} \mathcal{T}_{j}\}$, $|A| = n$ and $g$ is some function to combine the $n$ atomic tasks. We discuss more details for $g$ in Appendix~\ref{appx:prelim_details}.

\textbf{Compositional reasoning from atomic CoT.} We assume access to atomic CoT data $D^{\mathrm{CoT}}_{A} = \{D^{\mathrm{CoT}}_{\mathcal{T}_{i}} | \mathcal{T}_{i} \in A \} $. Models fine-tuned on a subset of $D^{\mathrm{CoT}}_{A}$ are \textbf{atomic CoT models}.

For their composition $\mathcal{T}_{A}$, we only have access to a training dataset $D_{\mathcal{T}_{A}}$ of size $N_{\mathcal{T}_{A}}$. We make two data assumptions following considerations about how compositional data would work in practice: (1) \textbf{Answer only:} The data only contains the answers as labels and \emph{not} labeled CoT traces. This reflects that high-quality annotated CoT supervision may be harder to obtain in practice than correct answers; (2) \textbf{Limited compositional supervision:} We assume $N_{\mathcal{T}_{A}}$ is small. We may be able to collect a small amount of data for each new compositional task of interest, but there are too many possible compositional tasks to collect large-scale data on.

\section{Composable Chains-of-Thought}
\label{sec:method}

We assume the CoT traces in each of the $n$ atomic task datasets follow a certain distribution distinct to that dataset. A pre-trained LLM $M_{0}$ fine-tuned on the atomic CoT data can be seen as a mixture model: it can generate CoT traces from each of those $n$ distributions, but it is unclear whether such models can produce compositional CoTs for compositional tasks. We observe that without additional supervision signals, such fine-tuned models typically only replicate one of the learned atomic reasoning patterns in the generated CoT; we show empirical evidence in Section~\ref{sec:intrinsic_eval}.

To compose $n$ atomic CoTs in one sequence $\mathbf{t}_{1}...\mathbf{t}_{n-1}\mathbf{t}_{n}$, the model must allocate substantial probability to $p(\mathbf{t}_{1}...\mathbf{t}_{n-1}\mathbf{t}_{n} \mid \mathbf{q})$, when the model is never trained on a sequence of CoTs. Our goal is to augment the atomic CoT training data $(\mathbf{q},\mathbf{t}, a)$ into  $(\mathbf{q},\mathrm{proxy\,prefixes},\mathbf{t},a)$, such that \textbf{the training data looks more in-distribution to the compositional data while not explicitly training models on compositional examples.}

\subsection{Composable CoT Training Data}
\label{sec:ccot_construction}

\begin{figure}
\begin{center}
  \centering
  
      \includegraphics[width=\linewidth,trim=42mm 120mm 120mm 61mm,clip]{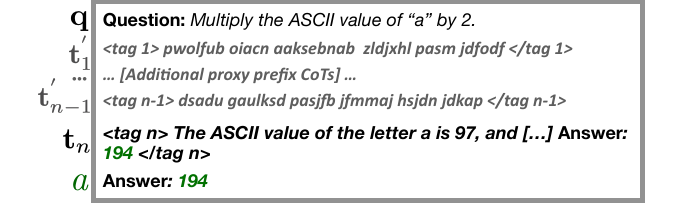}
  
\end{center}
\footnotesize
\caption{\label{fig:cot_construction} Construction of Composable CoT data. We insert $n-1$ \emph{proxy prefixes}, shown here as sequences of randomly sampled letters, at the end of the prompt, before the CoT.}
\end{figure}

Consider an atomic CoT dataset $D^{\mathrm{CoT}}_{\mathcal{T}} = \{(\mathbf{q},\mathbf{t},a)\}$ for $\mathcal{T} \in A$; we call this \textbf{standard CoT} data. We augment it with a set of \emph{chain-of-thought tags} $\mathcal{P}=\{p_{k}\}$ for $k \in \{1,...,n\}$.

\textbf{Proxy Prefixes.} Our goal is to augment standard CoT data such that atomic CoT models can learn a proxy distribution that simulates the distribution of the composition of atomic skills, despite not seeing compositional data. Thus, we append proxy prefixes to the prompt to simulate conditional generation of a CoT given other CoTs. In principle, these sampled strings should closely resemble CoT traces $\mathbf{t}$ that will show up in the test distribution. We do not assume knowledge of specific task compositions needed at inference time, but we can imagine sampling reasoning traces for problems related to the atomic task in question, or sampling reasoning-like strings from pre-training data.

Surprisingly, the approach we found to perform best is to use random-length sequences of randomly sampled letters as proxy prefixes.  Appendix~\ref{appx:proxy_prefix_cot} shows comparison to more realistic variants. One advantage of random letters is that they do not introduce bias towards any particular distribution of reasoning modes or related tasks, suggesting that learning the structure of Composable CoT is what drives generalization in our settings.

\textbf{Data Construction.} We sample a value of $k$ for each training example $d=(\mathbf{q},\mathbf{t},a)$, and we treat $\mathbf{t}$ as the $k$-th step in a notional compositional reasoning process. To achieve this, we append $k-1$ proxy prefixes $(\mathbf{t}^{'}_1\ldots\mathbf{t}^{'}_{k-1})$ to the end of the prompt: $\mathbf{t}^{'}_{i} = $\emph{<tag i>}$\mathbf{t}_{i}$\emph{</tag i>} for $1\leq i \leq k-1$ and $t_{i}$ is the $i$-th proxy prefix. By doing so, we obtain the augmented example $d^{'}=(\mathbf{q}\mathbf{t}^{'}_{1}\ldots\mathbf{t}^{'}_{k-1},\mathbf{t}_k)$ where $\mathbf{t}_k = $\emph{<tag k>}$\mathbf{t}\,a$\emph{</tag k>}.

Figure~\ref{fig:cot_construction} illustrates the procedure when $k=n$. The standard CoT $\mathbf{t}$ is: ``\emph{The ASCII value of the letter a is 97, and [...].}'' We augment the example by: (1) Appending $n-1$ proxy prefixes to the end of the question $\mathbf{q}$ to obtain the augmented prompt $\mathbf{q}\mathbf{t}^{'}_1\ldots\mathbf{t}^{'}_{n-1}$, with each proxy prefix wrapped in a tag; (2) Wrapping the CoT and the answer in a different tag \emph{<tag n>} as $\mathbf{t}_{n}$.

We use the scheme above to augment each example in the standard CoT dataset and obtain the augmented dataset $D^{\mathrm{aug}}_{\mathcal{T}}$. At inference time, we do not know a given atomic CoT will be used in which part of the compositional reasoning trace. Because CoT traces in $D^{\mathrm{aug}}_{\mathcal{T}}$ can simulate any of the $k$-th positions, models trained on $D^{\mathrm{aug}}_{\mathcal{T}}$ should be compatible with compositions of \emph{arbitrary order} instead of priming to any particular order seen during training. 

\textbf{Learning Objective.} Then, we fine-tune $M_{0}$ on $D^{\mathrm{aug}}_{\mathcal{T}}$ with a supervised fine-tuning objective:

$\mathcal{L}_ {D^{\mathrm{aug}}_{\mathcal{T}}}(\theta) = \frac{1}{N}\sum_{d^{'} \in D^{\mathrm{aug}}_{\mathcal{T}} } \mathcal{L}_ {d^{'}}(\theta)$ where $\mathcal{L}_ {d^{'}}(\theta) = -\log p_\theta(\mathbf{t}_{k} \mid \mathbf{q}...\mathbf{t}^{'}_{k-1})$. In other words, for each augmented example, we minimize the negative log likelihood of generating the CoT and answer, conditioned on the question and the $(k-1)$ proxy prefixes.

Note that when $k=1$, $d^{'}$ does not have any proxy prefix in the augmented prompt, so the model learns to generate CoT traces conditioned only on the question on those examples (e.g., the top right example in Figure~\ref{fig:composable_cot}). This simulates the scenario where an atomic CoT serves as the initial step of the compositional reasoning. For $1 < k \leq n$, the model learns to generate CoT conditioned on both the question and proxy prefixes (e.g., the bottom right example in Figure~\ref{fig:composable_cot}).

\textbf{Instantiation of Tags.} In practice, models only need to learn differentiations between the $n$-th tag, which marks the end of the notional $n$-way compositional reasoning, and all the other tags, which mark intermediate steps. Thus, we set $p_{n} = $\emph{<suffix>}, and all other $(n-1)$ tags as \emph{<prefix>}. 
Despite only having two instantiations of the tag, this scheme supports training of any length of $n$ of compositional CoT.

The scheme can also generalize to $n$-way composition \emph{at inference time}. Specifically, for $n > 2$, we 
can generate a CoT, then append the \emph{<suffix>} tag, continue to generate, and repeat $(n-1)$ times, thereby 
achieving test-time generalization to $n$-way composition. Details can be found in Section~\ref{sec:gen_beyond_pairwise}.

\subsection{Combining Atomic CoT Models}
After training an atomic CoT model on a single task $\mathcal{T}$, we combine multiple atomic CoT models to perform compositions. We consider two methods.

\textbf{ComposableCoT-MTL.} We apply multitask learning (MTL) to fine-tune $M_{0}$ on the combined dataset of $D^{\mathrm{aug}}_{A}  = \sum^{\mathcal{T}_{i} \in A} D^{\mathrm{aug}}_{\mathcal{T}_{i}}$ and obtain a single MTL model $M_{\mathrm{comb}}$ that learns all atomic tasks.

\textbf{ComposableCoT-Merge.} Model merging is another way to combine multiple models into a single multi-task model \citep{merging-fisher-weighting,yadav2023tiesmerging}. For each $\mathcal{T}_{i} \in A$, we start from $M_{0}$ and fine-tune a model $M_{i}$ (parametrized by $\theta_{i}$) on $D^\mathrm{aug}_{\mathcal{T}_{i}}$. Then we use Task Arithmetic \citep{task-arithmetic} to merge the $n$ models into a single model $M_{\mathrm{comb}}$ parametrized by $\theta_{\mathrm{comb}}$ as a linear combination of the deltas between each fine-tuned model parameter and the base model parameter: $\theta_{\mathrm{comb}} = \theta_{0} + \sum^{\mathcal{T}_{i} \in A} \alpha_{i}(\theta_{i} - \theta_{0})$ where $\alpha$ is the scaling factor.

\textbf{Inference.} 
When running zero-shot inference on the compositional task, we append \emph{<tag 1>} to the end of the prompt and sample a response from $M_{\mathrm{comb}}$. Then, we append the next tag to the end of the generated response, continue generation, and repeat by appending tags up to \emph{<tag n>}, as we assume that the number of compositions at test time is known.

\begin{algorithm}[]
\label{alg:rft}
\footnotesize
\caption{Rejection Sampling Fine-tuning on Composable CoT Models}\label{alg:cap}

\begin{algorithmic}[1]
\Require The combined model $M_{\mathrm{comb}}$; dataset $D_{\mathcal{T}_{A}} = \{(\mathbf{q}_{v},a_{v})\}^{N_{A}}_{v=1}$; the number of iterations $c$. 

\Ensure
\State $M_{0} \leftarrow M_{\mathrm{comb}}$ \Comment{Initialization} 

\For{$w$ in $1...c$}
\If{use rationalization} \Comment{Rationalization}
    \State $(\tilde{\mathbf{t}}_{v},\tilde{a}_{v})\leftarrow M_{w-1}(q_{v}a_{v})$ $\forall v \in \{1,...,{N_{A}}\}$ 
\Else
\State $(\tilde{\mathbf{t}}_{v},\tilde{a}_{v})\leftarrow M_{w-1}(q_{v})$ $\forall v \in \{1,...,{N_{A}}\}$
\EndIf

\State $D_{\mathrm{RFT}} \leftarrow \{(\mathbf{q}_{v},\tilde{\mathbf{t}}_{v},a_{v}) \ \mathrm{\ s.t.} \ v \in \{1,...,{N_{A}}\}\ \mathrm{and}$
\Statex \hspace{\algorithmicindent} $ \tilde{a}_{v} = {a}_{v}\}$ \Comment{CoTs with correct answers}
\State $M_{w} \leftarrow \mathrm{SFT}(M_{0},D_{\mathrm{RFT}})$ \Comment{Fine-tune the combined model on the accepted CoT data}
\EndFor
\end{algorithmic}
\end{algorithm}
\vspace{-0.1in}

\subsection{Improving Composition with Rejection Sampling Fine-tuning}
\label{sec:rft_alg}
$M_{\mathrm{comb}}$ can be
further improved with self-taught reasoning \citep{zelikman2022star} by rejection sampling fine-tuning (RFT) \citep{dong2023raft,yuan2024scaling}. Recall that for the compositional task, we only have answer labels instead of CoT traces. 
$M_{\mathrm{comb}}$ can serve as a starting point for RFT where we fine-tune $M_{\mathrm{comb}}$ with its own correct CoT responses using the limited compositional data.

Algorithm~\ref{alg:cap} shows the algorithm. Concretely, we sample responses from $M_{\mathrm{comb}}$ for each example in the compositional training data. Using the direct answer labels to verify the sampled responses, we can collect a supervised fine-tuning dataset $D_{\mathrm{RFT}}$ to continued fine-tune $M_{\mathrm{comb}}$. Such a process can be repeated for multiple iterations. For open-ended generation tasks that are hard to verify the correctness of sampled outputs only based on answer labels, we follow \citet{zelikman2022star,Ye-Durrett:2022:Fewshot} to perform rationalization to obtain $D_{\mathrm{RFT}}$; details can be found in Appendix~\ref{appx:rationalization}.

\begin{figure*}[t!]
\begin{center}
  \centering
  
   \includegraphics[width=\textwidth,trim=0mm 142mm 40mm 30mm]{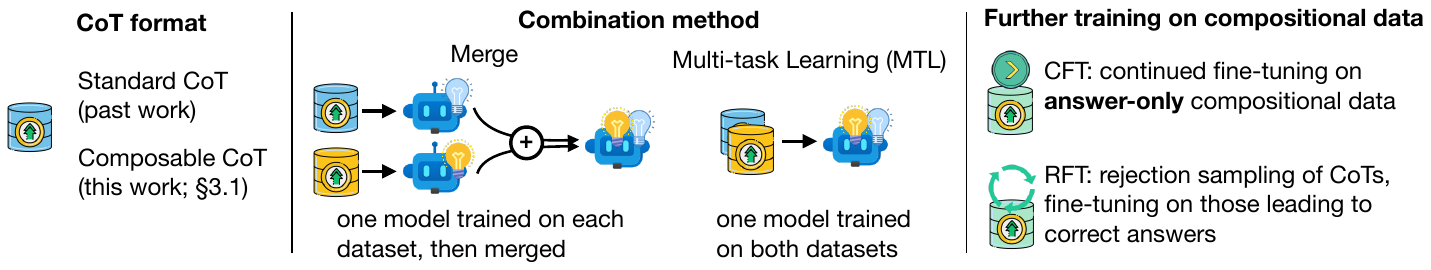}
  
\end{center}
\footnotesize
\caption{\label{fig:system_fig} Summary of settings for methods evaluated. Names in the results table reference configurations described in this figure; e.g., ComposableCoT-Merge uses ComposableCoTs with model merging, and in the zero-shot setting does not use further tuning.}
\end{figure*}

\begin{table*}[]
\footnotesize

\renewcommand{\arraystretch}{.8}
\centering

\begin{tabular}{lccccc}

\toprule
  \multirow{3}{*}{Methods} & \multicolumn{1}{l}{Next Letter} & \multicolumn{1}{l}{Concat}& \multicolumn{1}{l}{Concat} & \multicolumn{2}{c}{Skill-Mix Literary} \\
    & \multicolumn{1}{l}{+ Mult} & \multicolumn{1}{l}{+ Next Letter} & \multicolumn{1}{l}{+ Mult} & \multicolumn{2}{c}{+ Rhetorical} \\
   & \multicolumn{1}{c}{EM} & \multicolumn{1}{c}{EM} & \multicolumn{1}{c}{EM} & Full Marks &  Skill Fraction \\
   \midrule
   \midrule
 \multicolumn{6}{c}{Llama 2-7B} \\
 \midrule
 \midrule

 Few-shot Answer & 1.0 & {0.0} & 0.0 & 4.1   & 16.4 \\
 Few-shot CoT & 2.0 & {3.0} & 1.0 &  7.3 & 23.1 \\
\midrule
 StandardCoT-Merge& 2.0 & {12.5} & 2.3 & {11.0}  & {31.6} \\
 ComposableCoT-Merge (Ours) & 16.0 & \textbf{19.1} & 3.0 & 19.6 & {37.1} \\
  \midrule
StandardCoT-MTL & 5.0 & {0.0} & 0.0 & {17.6}  & {38.7} \\
ComposableCoT-MTL (Ours) & \textbf{18.7}  & 6.5 & \textbf{3.1}  & \textbf{22.9}  & \textbf{49.9}  \\
 \midrule
 \midrule
 \multicolumn{6}{c}{Qwen 2.5-7B} \\
 \midrule
 \midrule
 
 Few-shot Answer & 2.4 & 0.0 &2.7 & 34.7   & 56.0 \\
 Few-shot CoT & 2.0 & 0.0  & 21.3& 31.8   & 41.6  \\
 \midrule
 StandardCoT-Merge & 70.4 & 54.8 & \textbf{77.0} & 29.8& 48.0 \\
 ComposableCoT-Merge (Ours) & 95.4 & 19.2 & 75.4 & 39.6 & 62.1 \\
 \midrule
  StandardCoT-MTL & 3.6 & 60.9 &	72.1	&42.0&	58.2 \\
 ComposableCoT-MTL (Ours)&\textbf{96.3}&	\textbf{63.3}&	74.3&	\textbf{49.0}&	\textbf{66.7}  \\
 \bottomrule
\end{tabular}
\caption{Zero-shot compositional generalization of ComposableCoT with different combination approaches vs.~baselines. \emph{Without any compositional supervision}, using model merging or multitask learning to combine atomic CoT models trained on Composable CoT data outperforms baselines across settings and models.
}

\label{tab:zero-shot}
\end{table*}
\section{Experimental Setup}
\label{sec:exp_setup}

We select controlled evaluation tasks with the following criteria: (1) \textbf{Atomic tasks reflect core LLM reasoning skills}: We select atomic tasks that are representative of core skills that span logical, arithmetic, and writing. Prior work \citep{wei2022chain,dziri2023faith,yu2024skillmix} has shown that these skills can reflect more complicated capabilities such as advanced math reasoning and creative writing; (2) \textbf{Atomic skills are distinguishable}: To ensure controlled experiments of compositional generalization, atomic skills need to be distinguished from each other so that learning one skill affects learning another skill as little as possible; (3) \textbf{Compositions are unseen during pretraining}: General reasoning tasks such as math word problems feature examples that are common in pretraining. Our tasks are less observed, allowing us to attribute the task performance to the efficacy of training approaches rather than better recall of pretraining data.

Our tasks involve string manipulation, arithmetic, and natural language skill composition. Each setting involves atomic tasks and compositional tasks. We ensure that all atomic tasks are learnable through supervised fine-tuning with a small amount of training data ($N_{\mathcal{T}}\leq 500$) as shown in Appendix~\ref{appx:single_task}. 
We also confirm that the selected compositional tasks are less frequently seen for pre-trained LLMs: Appendix~\ref{appx:base_model_metrics_on_comp} shows the high perplexity of the task datasets, and Table~\ref{tab:zero-shot} shows the low accuracy of few-shot prompting.

\begin{table*}[]
\renewcommand{\tabcolsep}{1.3mm}
\renewcommand{\arraystretch}{1.0}
\scriptsize
\centering
\begin{tabular}{llccccc}
\toprule
  \multirow{3}{*}{Category} &  \multirow{3}{*}{Method} & \multicolumn{1}{l}{Next Letter} & \multicolumn{1}{c}{Concat}& \multicolumn{1}{c}{Concat} & \multicolumn{2}{c}{Skill-Mix Literary} \\
    & & \multicolumn{1}{l}{+ Mult} & \multicolumn{1}{l}{+ Next Letter} & \multicolumn{1}{l}{+ Mult} & \multicolumn{2}{c}{+ Rhetorical} \\
   & & \multicolumn{1}{c}{EM} & \multicolumn{1}{c}{EM} & \multicolumn{1}{c}{EM} & Full Marks &  Skill Fraction \\
    \midrule
     \midrule
 \multicolumn{7}{c}{Llama 2-7B} \\
 \midrule
  \midrule
\multirow{3}{*}{SFT} & SFT on Base Model & 3.1 & {5.0} & {9.0} & 35.5 & 60.1 \\
 & CFT on StandardCoT-Merge & 2.0 & {16.0} & {14.0} & 44.1 & 65.1 \\
 & CFT on StandardCoT-MTL & 3.0 & {26.0} & {11.0} & 38.0 & 62.1 \\
 \midrule
MTL & StandardCoT + Comp Answer & 5.0 & \textbf{46.0} & {13.3} & 22.9 & 45.5 \\
\midrule
\multirow{2}{*}{RFT} & StandardCoT-Merge  &  0.0 & 23.0   & 29.7  & 26.1 & 52.0   \\
 & ComposableCoT-Merge (Ours) & \textbf{72.0} &  46.0 &  \textbf{40.0} &  \textbf{45.3} &  \textbf{66.6} \\
  \midrule
 \multirow{2}{*}{RFT} & StandardCoT-MTL  &  10.1 & 0.0   & 0.0  & 38.8 & 58.1   \\
 & ComposableCoT-MTL (Ours) & 23.4 &  \textbf{69.6} &  8.3 &  44.1 &  63.5 \\

 \midrule
  \midrule
 \multicolumn{7}{c}{Qwen 2.5-7B} \\
 \midrule
  \midrule
\multirow{3}{*}{SFT} & SFT on Base Model & - & 31.9 & 2.0 & 35.5 & 60.3 \\
 & CFT on StandardCoT-Merge & - & 41.1	&9.3	&51.0	&71.4 \\
 & CFT on StandardCoT-MTL & - &  60.3	&12.7	&34.7	&56.3 \\
 \midrule
MTL & StandardCoT + Comp Answer & - & 65.1&	7.1	&41.2	&55.3\\
\midrule
\multirow{2}{*}{RFT} & StandardCoT-Merge  & - & 83.3   & 83.6  & 51.8 & 69.0   \\
 & ComposableCoT-Merge (Ours) & - & 33.9 &  87.0 &  56.7 &  70.1  \\
 \midrule
 \multirow{2}{*}{RFT} & StandardCoT-MTL  & - &  82.1 & \textbf{89.0}  & 44.9 & 63.4  \\
 & ComposableCoT-MTL (Ours) & - & \textbf{86.9}	&88.4&	\textbf{57.6}	&\textbf{71.5} \\
\bottomrule
\end{tabular}
\caption{Compositional task performance of rejection sampling fine-tuning (RFT) upon merged Composable atomic CoT models and other baselines. \emph{Mult} stands for ASCII multiplication and \emph{concat} stands for letter concatenation. \emph{SFT} stands for supervised fine-tuning with the compositional answer data; \emph{CFT} stands for continued fine-tuning; \emph{MTL} stands for multitask learning method. Results on Next Letter + Mult are omitted because the zero-shot performance saturates. For Qwen2.5-7B, RFT on ComposableCoT variants achieves the best compositional performance using the same compositional answer data.}

\label{tab:with_supervision}
\end{table*}
\textbf{String manipulation and arithmetic tasks.} We consider the following atomic tasks. \textbf{(1) Next letter in alphabet} \citep{efrat-etal-2023-lmentry,edman-etal-2024-cute}; \textbf{(2) Letter concatenation} \citep{wei2022chain,zhou2023leasttomost}; \textbf{(3) ASCII multiplication} \citep{dziri2023faith,gambardella-etal-2024-language}. We consider the following compositions of two of the atomic tasks: (1) \textbf{Next letter + multiplication}; (2) \textbf{Concatenation + next letter}; (3) \textbf{Concatenation + multiplication}. Details of tasks and data can be found in Appendix~\ref{appx:synth_construct}.

\textbf{Natural language skills.} We adapt the compositional benchmark Skill-Mix \citep{yu2024skillmix}: Given the definition and an example of a language skill (e.g. hyperbole), the model writes a sentence to demonstrate the skill about a given topic. 
We consider an atomic task to be handling skills over a \emph{category} of skills, and we evaluate on two categories: literary devices (\emph{Literary}) and rhetorical devices (\emph{Rhetorical}). Atomic CoT traces for Skill-Mix are distilled from GPT-4o \citep{openai2024gpt4ocard}, following \citet{zhao2024can}. The composition tasks require writing a sentence with two provided skills sampled from the \textbf{literary} and \textbf{rhetorical} categories. Details can be found in Appendix~\ref{appx:skillmix_details}.

\textbf{Evaluation Metrics.} For Skill-Mix tasks, we use quality measure metrics for the generated sentence from \citet{yu2024skillmix} (namely, \emph{Full Marks} and \emph{Skill Fraction}) based on a rubric, and use GPT-4o-mini as a judge. Details can be found in Appendix~\ref{appx:skillmix_metrics}. All other tasks are evaluated using \emph{exact match} accuracy with a regex-based answer extractor to extract model answers from responses.

\textbf{Zero-shot/Few-shot Baselines.} Figure~\ref{fig:system_fig} summarizes the high-order variables of the configurations we evaluate. For zero-shot compositional generalization, we include the following baselines: (1) Few-shot direct answer prompting; (2) Few-shot CoT prompting; (3) Model merging of atomic CoT models (\emph{StandardCoT-Merge}); (4) Multitask learning of atomic CoTs (\emph{StandardCoT-MTL}).

\textbf{Baselines with Compositional Supervision.} We compare rejection sampling fine-tuning of Composable CoT models with the following baselines. (1) Continued fine-tuning (CFT) the multitask model of atomic CoTs (\emph{CFT on StandardCoT-MTL}); (2) Continued fine-tuning the merged model of atomic CoTs (\emph{CFT on StandardCoT-Merge}); (3) Multitask learning of atomic CoTs and compositional answers (\emph{StandardCoT + Comp Answer}).

Details of each baseline above can be found in Appendix~\ref{appx:baseline}. The differences of methods we evaluate for each setting are summarized in Table~\ref{tab:method_summary}.

\textbf{Data Construction.} For two-way compositions, we sample uniformly from 2 chain-of-thought tags, \emph{<prefix>} and \emph{<suffix>}, for data construction. At inference time, we first append \emph{<prefix>} to the prompt and sample from the combined model. Then, we append \emph{<suffix>} to the end of the generated response, and continue generation.

\textbf{Models and Training.} We use Llama 2 7B-base \citep{touvron2023llama2openfoundation} and Qwen2.5 7B-base \citep{qwen2025qwen25technicalreport} as models, and use LoRA \citep{hu2022lora} for supervised fine-tuning. Configuration details and hyperparameters are in Appendix~\ref{appx:hypers}.

\section{Results}
\subsection{Zero-shot Generalization}
\label{sec:zero_shot_results}

We evaluate the compositional generalization of ComposableCoT-Merge and ComposableCoT-MTL \emph{without access to compositional supervision}. For all methods that we compare with, we control the amount of training data to be the same.
Details of the training data can be found in Appendix~\ref{appx:statistics}.

\textbf{ComposableCoT achieves better zero-shot generalization.}
Table~\ref{tab:zero-shot} shows that ComposableCoT variants outperform baselines on a range of tasks for both models. Combining atomic CoT models trained on Composable CoT is better than combining models trained on standard CoT across settings, indicating better ``composability'' at inference time. Appendix~\ref{appx:two_way_with_distractors} shows results where Composable CoT models are trained on more skills than are needed for the target task; they outperform standard CoT models in this setting as well.

\subsection{Limited Compositional Supervision}
\label{sec:comp_results_with_supervision}

\begin{table}
\centering
\footnotesize
\label{tab:three_way_comp}
\renewcommand{\tabcolsep}{0.3mm}

\begin{tabular}{cccc}
\toprule
 & \multicolumn{1}{l}{String Tasks} & \multicolumn{2}{c}{Skill-Mix} \\
 \midrule
 & \multicolumn{1}{c}{EM} & \multicolumn{1}{c}{Full Mark} & \multicolumn{1}{c}{Skill Fraction} \\
 \midrule
StandardCoT-Merge & 61.3 & 13.1 & 42.7 \\
ComposableCoT-Merge & 63.1 & 19.2 & 54.1 \\
\midrule
StandardCoT-MTL & 82.3 & 28.2 & 55.9 \\
ComposableCoT-MTL & \textbf{86.7} & \textbf{33.1} & \textbf{61.0} \\
\bottomrule
\end{tabular}
\caption{Zero-shot three-way composition. Combining ComposableCoT models outperforms combining StandardCoT models on the composition of three tasks. }
\label{tab:three_way}

\end{table}

We evaluate Composable CoT models after being further improved with one iteration of RFT using the limited compositional supervision. We compare with multitask learning and continued fine-tuning baselines given the same compositional answer dataset $D_{\mathcal{T}_{(i,j)}}$ of size $N_{(i,j)} \leq 500$. 
Details of the data condition are in Appendix~\ref{appx:statistics}.

Table~\ref{tab:with_supervision} shows that with the same compositional training data, \textbf{using RFT on top of ComposableCoT-MTL and ComposableCoT-Merge achieves the best compositional task performance}, outperforming multitask learning and continued fine-tuning baselines across settings. 

Moreover, \textbf{RFT upon ComposableCoT models is generally better than RFT on StandardCoT models.} Using the same combination method (MTL or Model Merging), RFT upon ComposableCoT models outperforms the StandardCoT counterpart across models and tasks.

\begin{table}[]
\footnotesize
\centering

\renewcommand{\tabcolsep}{0.7mm}
\begin{tabular}{cccc}
\toprule
 & \multicolumn{1}{c}{Few-shot} & \multicolumn{1}{c}{ StandardCoT} & \multicolumn{1}{c}{ComposableCoT} \\
\midrule
\midrule
\multicolumn{4}{c}{\textit{Performance by Academic Subject}} \\ 
\midrule
\midrule
Algebra & 60.7 & 60.7 & \textbf{65.0} \\
Counting & 38.4 & 41.8 & \textbf{44.1} \\
Geometry & 30.5 & 37.0 & \textbf{39.7} \\
Inter. Algebra & 21.9 & 23.8 & \textbf{28.3} \\
Num. Theory & 38.0 & \textbf{41.3} & 41.1 \\
Prealgebra & 62.5 & \textbf{66.0} & 65.4 \\
Precalculus & 21.1 & \textbf{22.0} & 20.0 \\
\midrule
\textit{Overall} & 42.2 & 44.6 & \textbf{46.5} \\
\midrule
\midrule
\multicolumn{4}{c}{\textit{Performance by \# of Required Skills}} \\ 
\midrule
\midrule

$k=2$ & 56.0 & 58.2 & \textbf{59.1} \\
$k=3$ & 43.0 & 44.7 & \textbf{47.3} \\
$k=4$ & 32.0 & 34.5 & \textbf{37.4} \\
$k=5$ & 23.0 & \textbf{24.0} & 23.4 \\
\bottomrule
\end{tabular}
\caption{Skill composition of ComposableCoT on MATH with Qwen2.5-7B. When trained on atomic skills for math reasoning, ComposableCoT models outperform StandardCoT models by 1.9\% on average.
}
\label{tab:math_zero_shot}
\end{table}
\section{Results on Complex Compositions}

\subsection{Three-way Composition}
\label{sec:gen_beyond_pairwise}
We evaluate Composable CoT on zero-shot compositions of \emph{three} atomic tasks on Qwen2.5-7B using the following compositional tasks:  (1) Letter Concat + Next Letter + Mult (\emph{String Tasks}); (2) Skill-Mix Literary + Rhetorical + Logical (\emph{Skill-Mix}): Write a sentence using skills from each of the three Skill-Mix categories. We compare ComposableCoT models with StandardCoT models constructed by model merging or multi-task learning. Details are in Appendix~\ref{appx:details_three_way}.

Table~\ref{tab:three_way} shows that combining ComposableCoT models performs better on three-way composition: for example, ComposableCoT-MTL outperforms StandardCoT-MTL by $4.8\%$ on average.

\subsection{Skill Composition for Math Reasoning}
We evaluate ComposableCoT in a less controlled setting of math reasoning. We explore reasoning in the context of the MATH dataset \citep{hendrycks2021measuring}, where prior work shows that knowledge of different math subjects (skills in these subjects, e.g., ``solving equations'') must be applied to solve problems \citep{metacognitive,he2025skilltargetedadaptivetraining}.

\textbf{Setup.} We use the STAT dataset from \citet{he2025skilltargetedadaptivetraining} where each example is a math word problem annotated with a CoT trace and one subject-specific (e.g., algebra) math skill required to solve it. For each subject, we select the five most frequently used skills in the training set as the atomic skills, and take the corresponding training examples to form atomic CoT data. Note that each example is created to specifically tailor one skill, but this setting does not necessarily feature a clear separation of skills like our string manipulation tasks do; see further discussions in Appendix~\ref{appx:details_math}. Then, we fine-tune an atomic CoT model using Composable CoT format \emph{for each subject} with the same scheme described in Section~\ref{sec:exp_setup}. Finally, we evaluate the fine-tuned model on the test set of the corresponding subject in the MATH dataset, where each test example is considered a composition of atomic skills within that subject. 

\textbf{Training and Evaluation.} We use Qwen2.5-7B-Base as the model and use multi-task learning 
as the combination method. We use the same low-data setting from Section~\ref{sec:exp_setup} with $100-500$ training examples for each subject. We compare against few-shot CoT prompting and SFT on StandardCoT data, and we use exact match accuracy as the evaluation metric. More details are in Appendix~\ref{appx:details_math}.

\textbf{Results.} Table~\ref{tab:math_zero_shot} shows that \textbf{combining ComposableCoT models
outperforms combining StandardCoT models in multiple subjects in MATH}, with an average improvement of 4.3\% over the base model and 1.9\% over the StandardCoT model. We also show a breakdown by the number of required skills. ComposableCoT outperforms StandardCoT not only on $k=2$, but also on $k=3$ and $k=4$ by substantial margins.

\begin{table}
\footnotesize
\centering

\renewcommand{\tabcolsep}{0.5mm}
\begin{tabular}{llrrrr}
\toprule
 & CoT & \multicolumn{1}{l}{Perf.} & \multicolumn{1}{l}{\% $\mathcal{T}_{1}$} & \multicolumn{1}{l}{\% $\mathcal{T}_{2}$} & \multicolumn{1}{l}{\% Both} \\
\midrule
 Next Letter  & Standard & 3.6 & 0.0 & 100.0 & 0.0 \\
+ Mult & Composable & 96.3 & 98.9 & 100.0 & $^\dagger$98.9 \\

 \midrule
 Concat & Standard & 60.9 & 100.0 & 66.7 & 66.7 \\
 + Next Letter & Composable & 63.3 & 100.0 & 85.9 & $^\dagger$85.9 \\
  \midrule
Concat  & Standard & 72.1 & 99.7 & 32.1 & 32.1 \\
 + Mult & Composable & 74.3 & 100.0 & 83.1 & $^\dagger$81.3 \\
 \midrule
Literary  & Standard & 42.0 & 65.3 & 58.0 & 37.6 \\
+ Rhetorical & Composable & 49.0 & 64.5 & 65.7 & $^\dagger$42.0 \\
 \bottomrule
\end{tabular}
\caption{Quality of CoTs generated by ComposableCoT zero-shot. ``\% $\mathcal{T}_{1}$'' denotes the percentage of generated responses that use the CoT format of the first atomic task of the composition, and likewise for the second. 
$^\dagger$ denotes that the ComposableCoT method has a significantly higher  ``\% Both'' than the StandardCoT counterpart at the $0.01$ level using a paired bootstrap test. ``Perf.'' denotes the task performance. }
\label{tab:instrinsic_eval}

\end{table}
\section{Analysis}
\label{sec:intrinsic_eval}

We evaluate the intrinsic quality of model outputs by analyzing the occurrence of atomic CoT patterns in CoTs generated by ComposableCoT models for zero-shot composition. 
For the string manipulation and arithmetic tasks, we extract template-based patterns of each atomic CoT from the generated outputs of models evaluated on the compositional task. For Skill-Mix, we consider the CoT pattern of an atomic task to be used if the generated response explicitly mentions the skill corresponding to that atomic skill category. 

Table~\ref{tab:instrinsic_eval} shows results with models trained from Qwen 2.5-7B and combined with MTL. Details and more results are in Appendix~\ref{appx:intrinsic_eval_full}. \textbf{Combining ComposableCoT leads to higher presence of both atomic CoT patterns in the generated responses compared to StandardCoT.} Models trained with the Composable CoT use the combination of learned skills in some form more frequently than StandardCoT. Examples of generated CoTs are in Appendix~\ref{appx:error_analysis}.

\section{Related Work}

Compositional generalization has been considered a core capability for human-level reasoning models \citep{Piantadosi216composition,Werchan20158MonthOldIS,Fodor1988ConnectionismAC, Lake2023HumanlikeSG}, and has been considered difficult for LLMs because of the limitations of the transformer architecture and autoregressive generation \cite{thomm2024limits,dziri2023faith}.
Recent theoretical analyses show that the compositional reasoning capability of LLMs can be improved by generating CoT \citep{li2024chain,li2023dissecting}, but empirical results show that non-trivial effort needs to be put through prompt engineering \citep{chen-etal-2024-skills,gao2024metareasoninglargelanguage} or data selection \citep{khot2023decomposed,zhou2023leasttomost,levy-etal-2023-diverse,ye-etal-2023-complementary} to observe such improvements with CoT \citep{sprague2025to}, particularly in domains where compositional solutions to problems are crucial \citep{knowledge-cross-word,ye2025longprocbenchmarkinglongcontextlanguage}. Prior work has explored more principled approaches, but they usually rely on heuristics to determine data quality \citep{hase-etal-2024-unreasonable,curriculum-learning} or involve computationally intensive methods \citep{sun2024easytohard,conklin-etal-2021-meta}. Our method is inspired by methods for merging models of different capabilities \citep{wu2025unlockingefficientlongtoshortllm,ma2025cotvalvelengthcompressiblechainofthoughttuning,tam2024realisticevaluationmodelmerging}, and our work is the first to use model merging for compositional generalization with CoT. 

\section{Conclusion}

\label{sec:conclusion}
We propose Composable Chain-of-Thought, a data augmentation scheme to convert CoT data of atomic reasoning skills into a format that facilitates inference-time compositional generalization. Training atomic CoT models with Composable CoT and combining them with model merging or multitask learning leads to better zero-shot compositional reasoning performance than building models with the standard CoT format. Such a combined model can be further improved by a limited amount of compositional data with rejection sampling fine-tuning. 
Learning to reason with Composable CoT shows a promising approach to improve compositional reasoning in LLMs, and could be extended to build more efficient and robust large reasoning models.
\section*{Limitations}
Our work focuses on the \emph{systematicity} aspect of compositional generalization, the capability of applying known components in unseen combinations \citep{Fodor1988ConnectionismAC}. We consider the productivity and primitive application aspects, including length generalization \citep{anil2022exploring,zhou2024transformers}, out of scope for this work. We believe that the composable CoT technique could help in this setting, but we leave more rigorous development and experimentation for future work.

Furthermore, our experiments leverage small-scale datasets where both the atomic skills and compositional skills can be learned with a small amount of training data. Our focus here is on conducting controlled experiments, but we believe our methods and their principles can be scaled up to more complex settings, as empirical results on math skill compositions have demonstrated.

Finally, our exploration is limited to open-source language models operating on English-language datasets, due to the availability of models and data for the task settings at hand. As additional models and datasets become available, we believe there would be value in scaling this study further to understand the results in a wider range of settings.

\section*{Ethical Considerations}
This work does not involve human subjects or the release of sensitive data. We do not clearly see harms arising from the applications of the proposed method. Therefore, we are not aware of any obvious ethical concern related to this work, except for those inherited from the development of large language models and generative AI technology more broadly.

\section*{Acknowledgments}

Thanks to Katrin Erk, Kyle Mahowald, and other members of the TAUR lab for helpful discussion and suggestions. This work was partially supported by the Sloan Foundation, a grant from Open Philanthropy, NSF CAREER Award IIS-2145280, the NSF AI Institute for Foundations of Machine Learning (IFML), and the NSF under Cooperative Agreement 2421782 and the Simons Foundation grant MPS-AI-00010515 awarded to the NSF-Simons AI Institute for Cosmic Origins — CosmicAI, https://www.cosmicai.org/. This research has been supported by computing support on the Vista GPU Cluster through the Center for Generative AI (CGAI) and the Texas Advanced Computing Center (TACC) at the University of Texas at Austin.

\bibliography{custom}

\appendix

\section{A Note on Composing Tasks}
 There exist various possible ways to combine atomic tasks into a compositional task with the combination function $g$. We simplify $g$ into two types: (1) composite: the output of one atomic task is used as part of the input of another task, $g(\mathcal{T}_{i}, \mathcal{T}_{j}) = \mathcal{T}_{i} \circ \mathcal{T}_{j}$ or $g(\mathcal{T}_{i}, \mathcal{T}_{j}) = \mathcal{T}_{j} \circ \mathcal{T}_{i}$; (2) concatenation: the outputs of the two atomic tasks are concatenated using the same input, $g(\mathcal{T}_{i}, \mathcal{T}_{j}) = \mathcal{T}_{i} \oplus \mathcal{T}_{j}$ or $g(\mathcal{T}_{i}, \mathcal{T}_{j}) = \mathcal{T}_{j} \oplus \mathcal{T}_{i}$. Among tasks evaluated in Section~\ref{sec:exp_setup}, the string manipulation and arithmetic tasks need to be solved by a composite function, while the Skill-Mix task can be solved by either a composite function or a concatenation function.
 
\label{appx:prelim_details}
\section{Design Choices for Proxy Prefixes}
\label{appx:proxy_prefix_cot}
When designing the proxy prefix CoT, we would like to consider the following constraints. (1) We do not assume any prior knowledge about what would possibly be put in the proxy prefix at inference time; 
(2) We do not assume strong relevance between the proxy prefix CoT and the actual CoT, i.e., not all the information in the proxy prefix CoT is useful for predicting the CoT and the final answer. Otherwise, the models might learn to copy the prefix CoT verbatim.

\subsection{Ablation of Proxy Prefix Variants on Compositions}
Based on the considerations above, we experiment with the following variants on the compositional setting: 

\begin{itemize}
 \item \textbf{Random Letters}: We uniformly sample random, lower-case letters from the alphabet to form a sequence 
 to simulate an \emph{arbitrary} prefix CoT. For each training example with a CoT trace of $m$ letters, the length of the prefix sequence is randomly drawn from a uniform distribution between $0.5m$ and $1.5m$, so that the random sequences have a similar length distribution as the atomic CoT traces.
 
\item \textbf{Scrambled CoT}: Given the ground truth CoT in an atomic CoT training example, we randomly shuffle the words in the ground truth CoT, combine them again into a sequence in a random order, and use the combined sequence as the prefix CoT. The tokens are in-distribution for the reasoning task, but the random ordering removes the structure typical of CoTs. Note that the ground truth CoT then appears twice: once scrambled, in the prefix, and again as the training target.    
\item \textbf{Random Text}: We sample random sentences from OpenWebText \citep{Gokaslan2019OpenWeb} and SmolLM \citep{allal2025smollm2smolgoesbig} to give prefix CoTs the same distribution as pre-training data.

\item \textbf{Random Text plus CoT}: We sample random sentences from the pre-training corpora and sentences from the ground truth CoT in the atomic CoT training example. We then concatenate these sentences in random order as the prefix CoT. Note that in this case, the model can simply copy from the proxy prefix to generate the target atomic reasoning.

\item \textbf{Random In-distribution (ID) CoT}: We randomly sample a ground truth atomic CoT from a different training example of the \emph{same} atomic task, and use it as the prefix. This removes the model's ability to copy the prefix CoT verbatim to do reasoning, while providing in-distribution CoT style and tokens.

\end{itemize}

We fine-tune Qwen2.5-7B on Composable CoT data for the string manipulation and arithmetic tasks using the prefix variants above. Then, we evaluate the fine-tuned models on zero-shot two-way compositions. Table~\ref{tab:proxy_prefix_composition} shows that using random letters as the prefixes leads to the best zero-shot compositional generalization among all variants. Interestingly, we note that the average performance generally decreases when the prefix distribution gets more similar to the atomic CoT distribution: while both use randomized sequences as prefixes, scrambled CoT underperforms random letters because scrambled CoT uses in-distribution words; while both use irrelevant sentences as prefixes, random ID CoT underperforms random text because random ID CoT uses in-distribution sentences. We hypothesize that this is because using in-distribution prefixes encourages models to copy the prefix CoT into the suffix CoT rather than generating a different CoT as the continuation, leading to less robust inference-time behaviors.

\subsection{Ablation of Proxy Prefix Variants on Atomic Tasks}

To further investigate the impact of different proxy prefix variants, we evaluate the following proxy prefix variants on atomic tasks, particularly to test the behavior under distribution shifts. We mainly investigate \textbf{random letters}, \textbf{random text}, and  \textbf{random text from the prompt}, where we sample random letters and words from the prompt $\textbf{q}$ to form a sequence of random lengths, to simulate three different distributions of prefix CoT.

We evaluate these variants by fine-tuning models on Composable CoT datasets \textbf{that only use the following augmentation}: $d^{'} =\mathbf{q}$ \emph{<prefix>} [proxy prefixes] \emph{</prefix>}  \emph{<suffix>} $\mathbf{t}a$ \emph{</suffix>}. Note that this is different from the setting discussed in Section~\ref{sec:ccot_construction} where the Composable CoT dataset consists of other possible augmentations as well based on the sampling of the tags (e.g., $d^{'} =\mathbf{q}$ \emph{<prefix>} $\mathbf{t}a$ \emph{</prefix>} when $k=1$). This experiment mainly aims at stress testing the model's capability of learning a single atomic task with a given proxy prefix CoT variant. We use the same hyperparameter configurations for all proxy prefix variants for a given task.

We evaluate the fine-tuned models on the in-domain task in two settings: (1) \emph{In-domain prefix}: we append the same type of prefix as we have used for training to the end of the prompt of the in-domain test example and evaluate the model on it; (2) \emph{Out-of-domain prefix}: we randomly sample a prefix from the other two variants and append it to the end of the prompt of the in-domain test example and evaluate the model on it. We run experiments on the three string manipulation and arithmetic tasks and report the average performance. Table~\ref{tab:proxy_prefix_atomic} shows that although using random letters as the proxy prefix leads to the worst in-domain performance, it generalizes the best to out-of-domain prefixes, which is a more desirable behavior.

\begin{table*}

\centering
\begin{tabular}{ccccc}
\toprule
 & Next Letter + Mult & Concat + Next Letter & Concat + Mult & \textit{Avg.} \\
 \midrule
Random Letters & \textbf{96.3} & \textbf{63.3} & \textbf{74.3} & \textbf{78.0} \\
Scrambled CoT & 93.3 & 59.3 & 73.6 & 75.4 \\
Random Text plus CoT & \multicolumn{1}{c}{63.3} & 61.5 & 57.0 & 60.6 \\
Random Text & 63.3 & 58.7 & 40.4 & 54.1 \\
Random ID CoT & 65.4 & 48.0 & 7.6 & 40.3 \\
\bottomrule
\end{tabular}
\caption{Zero-shot \textbf{composition} performance of Composable CoT models fine-tuned on different variants of proxy prefix on Qwen2.5-7B. Using random letters as the proxy prefix achieves the best performance.}
\label{tab:proxy_prefix_composition}
\end{table*}

\begin{table*}

\centering

\begin{tabular}{lrr}
\toprule
\multirow{2}{*}{Type of Proxy Prefix} & \multicolumn{2}{c}{Exact Match Accuracy} \\
 & \multicolumn{1}{l}{In Domain Prefix} & \multicolumn{1}{l}{Out-of-Domain Prefix} \\
 \midrule
Random Letters & 83.0 & 90.0 \\
Random Text from the Prompt & 86.4 & 82.5 \\
Random Text & 90.6 & 70.0 \\
\bottomrule
\end{tabular}
\caption{Atomic task performance of \textbf{atomic} CoT models fine-tuned on different variants of proxy prefix on Llama 2-7B. Using random letters as the proxy prefix achieves the best performance when evaluated with an unseen prefix at inference time.}
\label{tab:proxy_prefix_atomic}
\end{table*}

\section{Details of String Manipulation and Arithmetic Tasks}
\label{appx:synth_construct}
\subsection{Task Descriptions}
We consider the following atomic tasks.

(1) \textbf{Next Letter in Alphabet}: We adapt this task from \citet{efrat-etal-2023-lmentry,edman-etal-2024-cute}. Given a sequence of letters, find the next letter in the alphabet following the last letter.

(2) \textbf{Letter Concatenation}: We adapt this task from \citet{wei2022chain,zhou2023leasttomost}. Given a sequence of words, concatenate the first, second, last, or second-to-last letter of each word.

(3) \textbf{ASCII Multiplication}: We adapt this task from multi-digit multiplication tasks in \citet{dziri2023faith,gambardella-etal-2024-language}. Given a sequence of words and a letter, find the ASCII value of the given letter multiplied by a given constant value.

We consider the following pairwise composition of the atomic tasks.

(1) \textbf{Next Letter + Multiplication}: Given a sequence of letters, find the next letter in the alphabet following the last letter, determine its ASCII value, and then perform multiplication with a given constant.

(2) \textbf{Concatenation + Next Letter}: Given a sequence of words, concatenate the first, second, or second-to-last letter of each word and then find the next letter in the alphabet following the last letter of the concatenated sequence.

(3) \textbf{Concatenation + Multiplication}: Given a sequence of words, concatenate the first, second, or second-to-last letter of each word, find the ASCII value of the last letter of the concatenated sequence, and then perform multiplication.

\subsection{Data Construction}

\paragraph{Next Letter in Alphabet} We synthetically generate data for next letter in alphabet. We randomly sample letters from the English alphabet of a random length and concatenate them into a sequence. Then we extract the last letter from the sequence and derive the next letter following it in the alphabet. An example can be found in Example~\ref{example:atom:string-op:next-letter}. We automatically generate a chain-of-thought for each generated problem, using a fixed template shown in Example~\ref{example:atom:string-op:next-letter}.

\paragraph{ASCII Multiplication} Similarly, we randomly sample letters from the English alphabet of a random length and concatenate them into a sequence. Then, we randomly sample another letter $s$ and randomly sample an integer $a \in \{1,...,9\}$. We find the ASCII value of $s$ as $f(s)$ and compute the product $af(s)$ as the gold answer. An example can be found in Example~\ref{example:atom:string-op:ascii-multi}. We automatically generate a chain-of-thought for each generated problem, using a fixed template shown in Example~\ref{example:atom:string-op:ascii-multi}.

\paragraph{Letter Concatenation} We follow \citet{wei2022chain} to generate the dataset by randomly sampling from the most popular first and last names in the United States and the United Kingdom from \url{https://namecensus.com} and randomly concatenating them into a sequence of names. While the original task in \citet{wei2022chain} only requires concatenating the last letter of each name together, we raise the difficulty level by randomly asking for concatenations of the first, second, second-to-last, or the last letter. An example can be found in Example~\ref{example:atom:string-op:letter-concat}. The CoT template is also shown in Example~\ref{example:atom:string-op:letter-concat}.

\paragraph{Compositional Tasks} 
We synthetically construct the compositional tasks of the string manipulation and arithmetic tasks in similar procedures as used to generate the atomic data. An example of next letter + ASCII multiplication can be found in Example~\ref{example:comp:last+multi}, concatenation + next letter in Example~\ref{example:comp:concat+last}, and concatenation + multiplication in Example~\ref{example:comp:concat+multi}. We made a design decision to exclude one variant of concatenation + next letter that concatenates the last letter of each word and finds the next letter following the last letter in the concatenated sequence; this variant can be solved by the reasoning shortcut of only applying next letter in alphabet rather than a composition of both.

\begin{prompt}[title={\thetcbcounter{} Atomic Task Example: Letter Concatenation Example}, label=example:atom:string-op:letter-concat]
\char`[Instruction\char`]\\
Take the second-to-the-last letter of each word in the sequence and concatenate them in lower case: Tequan Monjur Khia Jodi-leigh answer\\

\char`[Chain-of-Thought + Answer String\char`]\\
The second-to-the-last letter of the 1st word is a. The second-to-the-last letter of the 2nd word is u. The second-to-the-last letter of the 3rd word is i. The second-to-the-last letter of the 4th word is g. So the answer is auig.\\

\char`[Answer String\char`]\\
auig

\end{prompt}

\begin{prompt}[title={\thetcbcounter{} Atomic Task Example: Next letter in alphabet}, label=example:atom:string-op:next-letter]

\char`[Instruction\char`]\\
Find the Next letter in alphabet following the last letter in the sequence: wqsisibnnicdlpwqbnoicdcxcxrfoilpcbnixuc\\bssssejxuzods answer: \\

\char`[Chain-of-Thought + Answer String\char`]\\
The last letter is s, and the letter following it in alphabet is t. So the answer is t.\\

\char`[Answer String\char`]\\
t

\end{prompt}

\begin{prompt}[title={\thetcbcounter{} Atomic Task Example: ASCII Multiplication}, label=example:atom:string-op:ascii-multi]

\char`[Instruction\char`]\\
Find the ASCII value of the letter after `<letter>' and multiply the ASCII value by 2: byaxaxcpoteznwnwseselyjlretx\\txcbfvmfezbycplymfotjbfv\\
jlhotzjbjcpycbtzhorepyjckofj <letter> d answer: \\

\char`[Chain-of-Thought + Answer String\char`]\\
The ASCII value of the letter d is 100, and multiplying the ASCII value by 2 gives us 200. So the answer is 200.\\

\char`[Answer String\char`]\\
200

\end{prompt}

\begin{prompt}[title={\thetcbcounter{} Compositional Task Example: Next letter + ASCII Multiplication}, label=example:comp:last+multi]

\char`[Instruction\char`]\\
Find the ASCII value of the Next letter in alphabet following the last letter in the sequence and multiply the ASCII value by 5: knnxqsxvshqugxfuquljumsbihgxvqihnxuu\\fuknxvumuupkpkshljqsbkiz answer: \\

\char`[Answer String\char`]\\
485

\end{prompt}

\begin{prompt}[title={\thetcbcounter{} Compositional Task Example: Concatenation + Next Letter}, label=example:comp:concat+last]

\char`[Instruction\char`]\\
Take the second-to-the-last letter of each word in the sequence, concatenate them in lower case, and find the Next letter in alphabet following the last letter in the sequence of the concatenated letters: Tyjai Ahijah Denzil Amine answer: \\

\char`[Answer String\char`]\\
o

\end{prompt}

\begin{prompt}[title={\thetcbcounter{} Compositional Task Example: Concatenation + Multiplication}, label=example:comp:concat+multi]

\char`[Instruction\char`]\\
Take the second-to-the-last letter of each word in the sequence, concatenate them in lower case, then find the ASCII value of the last letter in the sequence of the concatenated letters, and multiply the ASCII value by 3: Zarriah Amylee Li Javarie answer: \\

\char`[Answer String\char`]\\
315

\end{prompt}

\label{appx:synth_example}
\section{Details of Skill-Mix Tasks}
\label{appx:skillmix_details}
\subsection{Modifications of Skill-Mix}
We adapt the Skill-Mix dataset from \citet{yu2024skillmix}. For each example, the model is given a natural language skill, its definition, an example of the skill, and a topic to focus on, and the model needs to write a grammatical sentence to demonstrate the skill on the topic. Because we mainly focus on pairwise composition, we only use the $k=2$ and $k=1$ composition sets of the Skill-Mix data. We apply the following modifications to the dataset to fit our setting of compositional reasoning.
\begin{enumerate}
    \item Filtering the categories of skills: We keep examples with skills of the rhetorical and literary categories out of the five categories from the original dataset. This is because the rhetorical and literary skills have the least overlap  while other categories have more (e.g. the logical and rhetorical skills have a large body of overlaps).
    \item Removing the requirements of post-hoc explanation and refinement from the prompt. The original dataset evaluates models by prompting the models to first write a sentence, provide an explanation for the written sentence, and then do another round of refinement based on feedback from the grader (an LLM-as-a-judge). To fit into our setting of chain-of-thought reasoning and direct answering, we remove these irrelevant elements in the prompt.
    \item Using a public test set: The original evaluation of \citet{yu2024skillmix} hides the test set from the public and models can only be evaluated based on API calls to the hidden test set. To ensure reproducibility of our results, we use a public test set collected by \citet{zhao2024can}.
\end{enumerate}
As an open-ended generation task, Skill-Mix does not have a single ground truth sentence. \citet{zhao2024can} collect a silver-labeled dataset of sentences and their post-hoc explanations generated by GPT-4o \citep{openai2024gpt4ocard} and graded as correct by the grader. We further prompt GPT-4o to transform the (sentence, explanation) pairs into (CoT, sentence) pairs to resolve potential coreference issues. Examples of the prompt, CoT traces, and sentence can be found in Examples~\ref{example:atom:skill:literary}, ~\ref{example:atom:skill:rhetorical}, and ~\ref{example:comp:literary+rhetorical}.

\begin{prompt}[title={\thetcbcounter{} Atomic Task Example: Skill-Mix-Literary}, label=example:atom:skill:literary]

\char`[Instruction\char`]\\
Greetings! I am interested in natural language processing and I was wondering if you could help me generate an example of text that illustrates multiple skills in semantics or syntax. The example should be a minimal natural piece of text with up to a few lines in the context of Vikings that illustrates all of the following skills: oxymoron. Please keep the text as short as possible, and make sure the concepts can be found fully from the text. \\\\For reference, here are the definitions and examples for the concepts:\\***oxymoron**: A figure of speech in which apparently contradictory terms appear in conjunction. For example, ``The hall was filled with a deafening silence.'' Here, the phrase ``deafening silence'' is an oxymoron.\\\\Please start the minimal natural piece of text with `Answer:'.\\\\Thanks very much!\\

\char`[Chain-of-Thought + Answer String\char`]\\
Explanation: The goal is to construct a sentence that showcases the use of an oxymoron while remaining relevant to the Viking theme. To achieve this, we focus on creating a phrase that contains seemingly contradictory terms. We begin with the context of Vikings enduring harsh weather conditions. The word ``fiery'' conveys an intense heat, while ``frost'' indicates severe cold. When these terms are combined into ``fiery frost,'' they form an oxymoron as they depict both extreme temperature sensations simultaneously. This juxtaposition effectively creates a semantic paradox, leading to vivid imagery and emphasizing the harsh conditions faced by Vikings. \\\\Answer: ``In the bitter cold, the Viking felt a fiery frost.''\\

\char`[Answer String\char`]\\
Answer: ``In the bitter cold, the Viking felt a fiery frost.''

\end{prompt}

\begin{prompt}[title={\thetcbcounter{} Atomic Task Example: Skill-Mix-Rhetorical}, label=example:atom:skill:rhetorical]

\char`[Instruction\char`]\\
Greetings! I am interested in natural language processing and I was wondering if you could help me generate an example of text that illustrates multiple skills in semantics or syntax. The example should be a minimal natural piece of text with up to a few lines in the context of Hiking that illustrates all of the following skills: begging the question or assuming the conclusion. Please keep the text as short as possible, and make sure the concepts can be found fully from the text. \\\\For reference, here are the definitions and examples for the concepts:\\***begging the question or assuming the conclusion**: A circular argument that paraphrases the question. For example, ``Of course the earth is round; people have known this for hundreds of years.''\\\\Please start the minimal natural piece of text with `Answer:'.\\\\Thanks very much!\\

\char`[Chain-of-Thought + Answer String\char`]\\
Explanation: To illustrate the concept of begging the question or assuming the conclusion within the context of hiking, we start by identifying a statement that relies on its own conclusion as a premise. First, recognize that the claim ``hiking is beneficial'' needs supporting evidence to be considered valid. However, if we use the phrase ``because it's good for your health'' as justification, we notice that it merely reiterates the initial claim using slightly different wording. This is because declaring something ``beneficial'' inherently implies a positive impact, such as being ``good for your health.'' Thus, the reasoning becomes circular, as it depends on the same assumption it seeks to prove. \\\\Answer: ``Hiking is beneficial because it's good for your health.''\\

\char`[Answer String\char`]\\
Answer: ``Hiking is beneficial because it's good for your health.''

\end{prompt}

\begin{prompt}[title={\thetcbcounter{} Compositional Task Example: Skill-Mix Literary + Rhetorical}, label=example:comp:literary+rhetorical]

\char`[Instruction\char`]\\
Greetings! I am interested in natural language processing and I was wondering if you could help me generate an example of text that illustrates multiple skills in semantics or syntax. The example should be a minimal natural piece of text with up to a few lines in the context of Vikings that illustrates all of the following skills: anaphora resolution, begging the question or assuming the conclusion. Please keep the text as short as possible, and make sure the concepts can be found fully from the text. \\\\For reference, here are the definitions and examples for the concepts:\\***anaphora resolution**: Resolving the antecedent of a pronoun or noun phrase. For example, ``The car is falling apart, but it still works.'' Here , ``it'' is the anaphor and ``car'' is the antecedent.\\
***begging the question or assuming the conclusion**: A circular argument that paraphrases the question. For example, ``Of course the earth is round; people have known this for hundreds of years.''\\\\Please start the minimal natural piece of text with `Answer:'.\\\\Thanks very much!\\

\char`[Answer String\char`]\\
Answer:\\The Viking chief, undefeated thanks to his ship, asserted, ``It remains unconquered because it's the `Indomitable'.''

\end{prompt}

\subsection{Evaluation Metrics}
\label{appx:skillmix_metrics}
We use GPT-4o-mini as the LLM-as-a-judge to grade the generated sentence using the exact same grading rubric as provided by \citet{yu2024skillmix}; the grader judges the quality of the sentence based on if: (1) All skills are used; (2) The sentence makes sense; (3) The sentence attaches to the given topic; (4) The sentence is short. We use the evaluation metrics for each generated sentence in \citet{yu2024skillmix}, including the following:
\begin{enumerate}
\item \textbf{Full Marks:} $1$ if the generated sentence satisfies all four criteria above and $0$ otherwise.
\item \textbf{Skill Fraction:} The fraction of skills being demonstrated if all the other three criteria are satisfied; $0$ otherwise 
\end{enumerate}
We aggregate these metrics by averaging over all generated responses. In general, full marks evaluate the model's capability of writing a perfect sentence for the task, while skill fraction evaluates how good the model is at handling skills given that it is good at the other stylistic capabilities. We use Curator \citep{curator} for an efficient implementation of the evaluation pipeline.

\subsection{RFT on Skill-Mix Tasks with Rationalization}
\label{appx:rationalization}
For open-ended generation tasks like Skill-Mix, it is hard to only use the reference answer to verify the correctness of the responses sampled from a model. Thus, we use rationalization to perform rejection sampling fine-tuning for Skill-Mix:
we first append the direct answer label to the end of the prompt and sample post-hoc explanations for the given answer from the model; because $M_\mathrm{comb}$ is optimized to generate an answer following a CoT, we extract the generated answer following the generated explanation and filter out explanations whose following answer is not the same as the provided gold answer; finally, we use the accepted explanations as surrogates for CoT to form the RFT data.

\section{Details of Baselines}
\label{appx:baseline}

\textbf{Zero-shot/Few-shot Baselines.} We include the following baselines: (1) Few-shot direct answer prompting: we prompt $M_{0}$ with 5-shot demonstrations drawn from the compositional data; (2) Few-shot CoT prompting: we prompt $M_{0}$ with 5-shot CoT demonstrations drawn from the \emph{atomic} data; (3) Model merging of atomic CoT models (\emph{StandardCoT-Merge}): we fine-tune two models $M_{i}$ and $M_{j}$ based on $M_{0}$ with $D^\mathrm{CoT}_{\mathcal{T}_{i}}$ and $D^\mathrm{CoT}_{\mathcal{T}_{j}}$ respectively and merge them into $M_\mathrm{comb}$
with Task Arithmetic; (4) Multitask learning of atomic CoTs (\emph{StandardCoT-MTL}): we fine-tune $M_{0}$ to be a single multitask learning model $M_\mathrm{SCoT-MTL}$ on $D^\mathrm{CoT}_{\mathcal{T}_{i}} +  D^\mathrm{CoT}_{\mathcal{T}_{j}}$.

\textbf{Baselines with Compositional Supervision.} We use the following baselines: (1) Continued fine-tuning (CFT) the multitask model of atomic CoTs (\emph{CFT on StandardCoT-MTL}): we continue fine-tune the multitask model $M_\mathrm{SCoT-MTL}$ on $D_{\mathcal{T}_{(i,j)}}$; (2) Continued fine-tuning the merged model of atomic CoTs (\emph{CFT on StandardCoT-Merge}): we continue fine-tune the merged model of the two atomic CoT models $M_\mathrm{comb}$ on $D_{\mathcal{T}_{(i,j)}}$; (3) Multitask learning of atomic CoTs and compositional answers (\emph{StandardCoT + Comp Answer}): we fine-tune a single multitask learning model based on $M_{0}$ on the combined dataset of $D^\mathrm{CoT}_{\mathcal{T}_{i}} + D^\mathrm{CoT}_{\mathcal{T}_{j}} + D_{\mathcal{T}_{(i,j)}}$. We also include supervised learning baselines (SFT) where $M_{0}$ is fine-tuned on the same compositional answer data $D_{\mathcal{T}_{(i,j)}}$.

\section{Single-Task Learning Performance}
\label{appx:single_task}
We report the single-task learning performance of the atomic CoT models by evaluating them on the in-domain atomic tasks. We would like the atomic tasks to be easy to learn to reflect the practical settings where we train models on basic, easy-to-learn skills and generalize to harder, unseen tasks. The training data conditions and hyperparameters for training can be found in Appendix~\ref{appx:hypers}. Table~\ref{tab:single_task} shows that all atomic tasks we evaluate are learnable within a small amount of training data ($N_{i},N_{j} \leq 500)$. 

In addition, we observe that training on ComposableCoT or StandardCoT does not lead to consistent differences in atomic CoT performance, while the exception is on Skill-Mix-Rhetorical for Llama 2-7B where fine-tuning on ComposableCoT outperforms fine-tuning on StandardCoT by a large margin. 
\begin{table*}
\footnotesize

\centering
\renewcommand{\tabcolsep}{1.1mm}
\begin{tabular}{lccccccc}
\toprule
\multirow{2}{*}{CoT Format} & \multicolumn{1}{c}{Next Letter} & \multicolumn{1}{c}{ASCII Mult} & \multicolumn{1}{c}{Concat} & \multicolumn{2}{c}{Skill-Mix Literary} & \multicolumn{2}{c}{Skill-Mix Rhetorical} \\
 & \multicolumn{1}{c}{EM} & \multicolumn{1}{c}{EM} & \multicolumn{1}{c}{EM} & \multicolumn{1}{c}{Full Marks} & \multicolumn{1}{c}{Skill Fraction} & \multicolumn{1}{c}{Full Marks} & \multicolumn{1}{c}{Skill Fraction} \\
  \midrule
\multicolumn{8}{c}{Llama 2-7B} \\
 \midrule
StandardCoT & 100.0 & 85.7 & 83.0 & 63.5 & 63.5 & 53.3 & 53.3 \\
ComposableCoT & 95.0 & 86.0 & 77.0 & 71.4 & 71.4 & 72.4 & 72.4 \\
 \midrule
\multicolumn{8}{c}{Qwen 2.5-7B} \\
\midrule
StandardCoT & 90.0 & 99.0 & 77.4 & 77.6  & 77.6 & 70.5 & 70.5 \\
ComposableCoT & 99.4 & 99.7 & 77.3 & 77.6 & 77.6 & 76.7 & 76.7 \\
\bottomrule

\end{tabular}
\caption{Single-task learning performance by evaluating the atomic CoT models on the in-domain atomic tasks.}
\label{tab:single_task}
\end{table*}

\section{Base Model Performance on Evaluation Tasks}
\label{appx:base_model_metrics_on_comp}

To confirm that our task design includes evaluation tasks that are less commonly seen in the pretraining data of LLMs, we evaluate the perplexity of the task datasets.

We compare the datasets of the string manipulation and arithmetic datasets used in our experiments with mathematical reasoning data (GSM8K \citep{Cobbe2021TrainingVT}), and instruction following data (Alpaca \citep{alpaca}) in terms of perplexity: we compute the average perplexity score of pre-trained LLMs over the concatenation of the question and ground-truth chain-of-thought response to examine how predictable the task is; the lower the perplexity is, the more predictable and the easier the task is to learn. We also include the perplexity over the pretraining corpus as a reference point.
\begin{table*}

\footnotesize
\centering
\begin{tabular}{cccccc}
\toprule 
\multicolumn{2}{c}{\multirow{2}{*}{Task}} & \multirow{2}{*}{Pretraining} & Math & Instruction & String Manipulation \\
&  & & Reasoning & Following & And Arithmetic \\
\multicolumn{2}{c}{Dataset} & WikiText & GSM8K & Alpaca & Avg. \\
\midrule
\multirow{2}{*}{Model} & Llama 2-7B & \multicolumn{1}{c}{4.77} & \multicolumn{1}{c}{2.54} & \multicolumn{1}{c}{3.84} & \multicolumn{1}{c}{15.97} \\
 & Qwen 2.5-7B & \multicolumn{1}{c}{5.93} & \multicolumn{1}{c}{2.38} & \multicolumn{1}{c}{4.68} & \multicolumn{1}{c}{7.02} \\
 \bottomrule 
\end{tabular}
\caption{Perplexity of the base models over the task datasets. For the string manipulation and arithmetic tasks, the perplexity score is averaged over the 3 atomic tasks and the 3 pairwise compositional tasks. Our evaluation datasets include text that is less predictable under pre-trained LLMs than other similar tasks. }
\label{tab:perplexity}
\end{table*}

Table~\ref{tab:perplexity} indicates that our selected tasks consist of text that is less typical under pre-trained language models than other similar tasks, particularly other popular reasoning tasks. The higher perplexity is likely due to these tasks requiring the model to operate on letters rather than words.

\section{Training Configurations}
\label{appx:hypers}
\subsection{General Training Configurations}
We conduct all fine-tuning experiments with LoRA\citep{hu2022lora} using the following set of hyperparameters: we use a rank of 8, $\alpha=16$, and a dropout rate of 0.2 to prevent overfitting. We apply LoRA adapters to all linear modules, including the attention matrices $Q$, $K$, $V$ and MLP matrices of all layers. We use bfloat16 precision for training and we use the efficient implementation of LoRA by LlamaFactory \citep{zheng-etal-2024-llamafactory}. We use a training batch size of $4$ and train for $5$ epochs for all experiments that share the same number of training data; for methods that use a potentially smaller amount of training data (e.g. RFT methods usually get fewer data examples than the number of compositional training data provided, depending on how many correct responses we can sample from the model), we adjust the batch size to match the number of steps.

\subsection{General Inference Configuration}
For rejection sampling, we sample 10 responses for each prompt and use temperature $\tau=0.9$ for inference; otherwise, we use greedy decoding. 

\subsection{Configuration for Rejection Sampling Fine-tuning}
In addition to the sampling parameters (see Section~\ref{sec:exp_setup}), we consider the following configuration of RFT for sampling the correct responses: if the model generates multiple correct responses for a given question, we only randomly select \emph{one} of them to be added into the RFT dataset $D_{\mathrm{RFT}}$. In this way we ensure the diversity of examples in $D_{\mathrm{RFT}}$ so that the dataset will not be filled with samples from a small set of examples where the model is good at. For Skill-Mix tasks, we perform rationalization for RFT because it is open-ended generation (see Section~\ref{sec:rft_alg}).

\subsection{Hyperparameters: Learning Rate}
We find in preliminary experiments that learning rate is the most important hyperparameter for the fine-tuning experiments of our interest. We perform hyperparameter sweeps for learning rate over the space of $\{5e-3,1e-3,5e-4,1e-4,5e-5\}$ on a validation set for each experiment. The optimal learning rate for each method for the experiments with compositional supervision in Table~\ref{tab:lr_config}.
\begin{table*}
\renewcommand{\tabcolsep}{1.3mm}
\renewcommand{\arraystretch}{1.0}
\footnotesize
\centering
\begin{tabular}{llcccc}
\toprule
  \multirow{3}{*}{Category} &  \multirow{3}{*}{Method} & \multicolumn{1}{l}{Next Letter} & \multicolumn{1}{l}{Concat}& \multicolumn{1}{l}{Concat} & \multicolumn{1}{c}{Skill-Mix Literary} \\
    & & \multicolumn{1}{l}{+ Mult} & \multicolumn{1}{l}{+ Next Letter} & \multicolumn{1}{l}{+ Mult} & \multicolumn{1}{c}{+ Rhetorical} \\
    \midrule
     \midrule
 \multicolumn{6}{c}{Llama 2-7B} \\
 \midrule
  \midrule
\multirow{3}{*}{SFT} & SFT on Base Model & 1e-3 & 1e-3 & 5e-4 & 5e-4  \\
 & CFT on StandardCoT-Merge & 1e-3 & 5e-4 & 1e-4 & 1e-4 \\
 & CFT on StandardCoT-MTL & 1e-4 & 1e-4 & 1e-4 & 1e-3 \\
 \midrule
MTL & StandardCoT + Comp Answer & 1e-3 & 5e-4 & 1e-3 & 5e-4\\
\midrule
\multirow{2}{*}{RFT} & StandardCoT-Merge  &  - & 1e-3   & 1e-3  & 5e-4   \\
 & ComposableCoT-Merge (Ours) & 1e-4 &  1e-4 & 1e-3 &  1e-3  \\

 \midrule
  \midrule
 \multicolumn{6}{c}{Qwen 2.5-7B} \\
 \midrule
  \midrule
\multirow{3}{*}{SFT} & SFT on Base Model & - & 1e-3 & 1e-3 & 5e-4  \\
 & CFT on StandardCoT-Merge & - & 5e-4	&5e-4	&1e-4	 \\
 & CFT on StandardCoT-MTL & - &  1e-3	&1e-3	&1e-3\\
 \midrule
MTL & StandardCoT + Comp Answer & - & 5e-4&	5e-4	&1e-3	\\
 \midrule
 \multirow{2}{*}{RFT} & StandardCoT-MTL  & - &  1e-3 & 1e-4  & 5e-4   \\
 & ComposableCoT-MTL (Ours) & - & 1e-3	&1e-3&	5e-4	\\
\bottomrule
\end{tabular}
\caption{Optimal learning rate for each method in the experiments with compositional supervision.}
\label{tab:lr_config}

\end{table*}
\subsection{Hyperparameters: Model Merging}
For methods that use model merging as the combination, we use Task Arithmetic \citep{task-arithmetic} to combine the atomic CoT models. For two-way compositions, we perform a hyperparameter sweep for the scalars $\alpha_{1}$ over the space of $\alpha_{1} \in \{0.1,0.2,0.3,0.4,0.5,0.6,0.7,0.8,0.9\}$ on a validation set for each task. Another scaling factor is $\alpha_{2} = 1-\alpha_{1}$.

\section{Two-way Composition with Larger Skill Pools}
\label{appx:two_way_with_distractors}
\begin{table}
\centering
\footnotesize

\renewcommand{\tabcolsep}{0.4mm}
\begin{tabular}{cccc}
\toprule
 &  & \multicolumn{1}{l}{Standard} & \multicolumn{1}{l}{Composable} \\
 \midrule
Next Letter + Mult & EM & 56.1 & \textbf{75.9} \\
Concat + Next Letter & EM & 39.1 & \textbf{46.2} \\
Concat + Mult & EM & 44.3 & \textbf{48.9} \\
\midrule
Skill-Mix Literary & Full Mark & 37.1 & \textbf{42.0} \\
+ Rhetorical & Skill Fraction & 55.1 & \textbf{62.7} \\
\bottomrule
\end{tabular}
\caption{Zero-shot generalization on two-way compositions when merging \textbf{three} atomic models (i.e., there is a distractor skill). Merging ComposableCoT models is better than merging StandardCoT models in this setting.}
\label{tab:two_way_comp_merge_three}
\end{table}

In practice, models may need to have many capabilities to address problems of interest. Compared to our existing settings, such models need to select the skills to engage with for a particular task out of a larger pool of learned skills.

To evaluate this scenario, we train atomic models on \emph{three} atomic skills on Qwen 2.5-7B and evaluate the combined model (with model merging) on the zero-shot composition of \emph{two} of the atomic skills. This setup provides the model with additional potential combinations of skills to reason about. For the Skill-Mix tasks, we train on an additional atomic task, logical reasoning. Table ~\ref{tab:two_way_comp_merge_three} shows that learning ComposableCoT outperforms using StandardCoT by $8.8\%$ on average, indicating that ComposableCoT models can select the appropriate skills to compose out of many learned skills.

\section{Data Statistics}
\label{appx:statistics}
\subsection{General Data Conditions for Experiments}
Table~\ref{tab:data_condition_by_task} summarizes the amount of training data and test data used in the evaluations in Sections~\ref{sec:zero_shot_results} and ~\ref{sec:comp_results_with_supervision}. Note that for letter concatenation + multiplication we have two sizes of the compositional training data for Llama 2-7B and Qwen 2.5-7B: this is because all methods on Llama 2-7B perform poorly on zero-shot evaluation for this task and we need a slightly larger amount of compositional training data so that different methods can start to show distinguishable compositional task performance from each other. Regardless, we still consider $500$ to be a reasonably small amount of training data, satisfying our ideal data conditions defined earlier.

\begin{table*}
\centering
\begin{tabular}{llrr}
\toprule
 &  & \multicolumn{1}{l}{\# Train} & \multicolumn{1}{l}{\# Test} \\
 \midrule
\multirow{5}{*}{Atomic Tasks} & Next Letter & 100 & 700 \\
 & ASCII Mult & 100 & 700 \\
 & Concat & 500 & 700 \\
 & Skill-Mix Literary & 100 & 126 \\
 & Skill-Mix Rhetorical & 100 & 105 \\
  \midrule
\multirow{5}{*}{Compositional Tasks} & Next Letter + Mult & 100 & 700 \\
 & Concat + Next Letter & 100 & 504 \\
 & Concat + Mult (Llama 2-7B) & 500 & 700 \\
 & Concat + Mult (Qwen 2.5-7B) & 100 & 700 \\
 & Skill-Mix Literary + Rhetorical & 100 & 245 \\
 \bottomrule
\end{tabular}
\caption{Data conditions for each task used for our evaluation.}
\label{tab:data_condition_by_task}
\end{table*}

\subsection{Training Data Used by Each Method}
We show a detailed breakdown in Table~\ref{tab:data_condition_by_method_zeroshot} of the amount of training data used by each zero-shot method for both models and in Table~\ref{tab:data_condition_by_method} for Qwen 2.5-7B by each method with compositional answer data in the experiments in Section~\ref{sec:comp_results_with_supervision}. Note that the statistics for Llama 2-7B in the setting with compositional supervision are mostly the same except $N_{(i,j)} = 500$ for concat + next letter and concat + mult.

\begin{table*}
\centering
\begin{tabular}{llrr}
\toprule
 & Method & \multicolumn{1}{l}{$N_{i}$} & \multicolumn{1}{l}{$N_{j}$} \\
 \midrule
 & StandardCoT-Merge & 100 & 100 \\
Next Letter + Mult; & ComposableCoT-Merge & 100 & 100 \\
Skill-Mix Literary + Rhetorical & StandardCoT-MTL & 100 & 100 \\
 & ComposableCoT-MTL & 100 & 100 \\
 \midrule
& StandardCoT-Merge & 500 & 100 \\
Concat + Next Letter; & ComposableCoT-Merge & 500 & 100 \\
Concat + Mult & StandardCoT-MTL & 500 & 100 \\
 & ComposableCoT-MTL & 500 & 100 \\
 \bottomrule
\end{tabular}
\caption{The detailed breakdown of the number of training data used by each method in the zero-shot setting. $N_{i}$ and $N_{j}$ denote the number of training data from the atomic tasks $\mathcal{T}_{i}$ and $\mathcal{T}_{j}$ seen by the method during training.}
\label{tab:data_condition_by_method_zeroshot}
\end{table*}

\begin{table*}

\begin{tabular}{llrrr}
\toprule
 & Method & \multicolumn{1}{l}{$N_{i}$} & \multicolumn{1}{l}{$N_{j}$} & \multicolumn{1}{l}{$N_{(i,j)}$} \\
 \midrule
 & SFT on Base Model & 0 & 0 & 100 \\
 & CFT on StandardCoT-Merge & 100 & 100 & 100 \\
 & CFT on StandardCoT-MTL & 100 & 100 & 100 \\
Next Letter + Mult; & MTL on StandardCoT + Comp Answer & 100 & 100 & 100 \\
 Skill-Mix Literary + Rhetorical & RFT on StandardCoT-Merge & 100 & 100 & 100 \\
 & RFT on ComposableCoT-Merge & 100 & 100 & 100 \\
 & RFT on StandardCoT-MTL & 100 & 100 & 100 \\
 & RFT on ComposableCoT-MTL & 100 & 100 & 100 \\
  \midrule
 & SFT on Base Model & 0 & 0 & 100 \\
 & CFT on StandardCoT-Merge & 500 & 100 & 100 \\
 & CFT on StandardCoT-MTL & 500 & 100 & 100 \\
 & MTL on StandardCoT + Comp Answer & 500 & 100 & 100 \\
Concat + Next Letter; & RFT on StandardCoT-Merge & 500  & 100 & 100 \\
Concat + Mult & RFT on ComposableCoT-Merge & 500 & 100 & 100 \\
 & RFT on StandardCoT-MTL & 500 & 100 & 100 \\
 & RFT on ComposableCoT-MTL & 500 & 100 & 100 \\
 \bottomrule
\end{tabular}
\caption{The detailed breakdown of the number of training data used by each method with compositional supervision for Qwen 2.5-7B. $N_{i}$ and $N_{j}$ denotes the number of training data from the atomic tasks $\mathcal{T}_{i}$ and $\mathcal{T}_{j}$ seen by the method during training. $N_{(i,j)}$ denotes the number of compositional answer data seen during training.}
\label{tab:data_condition_by_method}
\end{table*}

\begin{table*}[]
\scriptsize

\centering
\begin{tabular}{llllll}

\toprule
\multirow{3}{*}{Method} & \# Atomic  & Atomic  & Combination &  Model  & How is \\
&  CoT Models &  CoT & Method  & trained on & Compositional Data \\
&Trained&Format&&Compositional Data &Used\\
\midrule
\midrule
\multicolumn{6}{c}{\emph{Zero-shot Evaluation}} \\
\midrule
\midrule
StandardCoT-Merge & 2 & Standard & Merge & - & - \\
\textbf{ComposableCoT-Merge (Ours)} & 2 & Composable & Merge & - & - \\
\midrule
StandardCoT-MTL & 1 & Standard & MTL & - & - \\
\textbf{ComposableCoT-MTL (Ours)} & 1 & Composable & MTL & - & - \\
\midrule
\midrule
\multicolumn{6}{c}{\emph{Evaluation with Limited Compositional Answer Data}} \\
\midrule
\midrule
CFT on StandardCoT-Merge & 2 & Standard & Merge & StandardCoT-Merge & CFT \\
CFT on StandardCoT-MTL & 1 & Standard & MTL & StandardCoT-MTL & CFT \\
\midrule
 \multirow{2}{*}{MTL on StandardCoT + Comp Answer} & \multirow{2}{*}{-} & \multirow{2}{*}{Standard} & \multirow{2}{*}{-} & \multirow{2}{*}{Pretrained Model} & Mix with Atomic\\
  &&&&&  CoT data and MTL\\
  \midrule
RFT on StandardCoT-Merge & 2 & Standard & Merge & StandardCoT-Merge & RFT \\
\textbf{RFT on ComposableCoT-Merge (Ours)} & 2 & Composable & Merge & ComposableCoT-Merge & RFT \\
RFT on StandardCoT-MTL & 1 & Standard & MTL & StandardCoT-MTL & RFT \\
\textbf{RFT on ComposableCoT-MTL (Ours)} & 1 & Composable & MTL & ComposableCoT-MTL & RFT \\
\bottomrule
\end{tabular}
\caption{Summary of methods evaluated in the zero-shot compositional evaluation and the composition with limited compositional answer data.``Merge'' stands for model merging; ``MTL'' stands for multitask learning; ``CFT'' stands for continued fine-tuning; ``RFT'' stands for rejection sampling fine-tuning. ``-'' means the property is not applicable to the method (e.g. \emph{MTL on Standard + Comp Answer} mixes Standard CoT data with compositional answer data, and trains a single MTL model from the pretrained model, so there is no atomic CoT model trained or combined.)  }
\label{tab:method_summary}
\end{table*}

\section{Details of Three-way Compositions}
\label{appx:details_three_way}
\subsection{Data}

We include 700 test examples for Letter Concat + Next Letter + Mult (\emph{String Tasks}), and 245 test examples for Skill-Mix Literary + Rhetorical + Logical (\emph{Skill-Mix}). For \emph{Skill-Mix}, we additionally train an atomic model for Skill-Mix-Logical with 100 training examples.

\subsection{Training and Inference}
\paragraph{Training} We use the same data augmentation scheme to create atomic CoT training data as the one we use for two-way composition in Section~\ref{sec:exp_setup}. This means that we append only one proxy prefix to the prompt. The general scheme can insert at most $n-1$ proxy prefixes at the end of the prompt for $n > 2$, but we found that the test-time generalization scheme described in \textbf{Instantiation of Tags} under Section~\ref{sec:ccot_construction} works as well: adding only one proxy prefix achieves comparable compositional performance to adding two proxy prefixes while being more efficient during training, since the training data length is shorter. Thus, we experiment with the latter scheme.

\paragraph{Inference} We use the same inference strategy specified in \textbf{Data Construction} under Section~\ref{sec:exp_setup}: for zero-shot inference, we first sample a response from $M_{\mathrm{comb}}$. Then, we repeat the following \emph{twice}: we append \emph{<suffix>} to the end of the generated response when it stops generation, and continue generation until the model stops again.

\section{Full Results for the Quality Analysis of the Generated CoTs}
\label{appx:intrinsic_eval_full}

\paragraph{Model Merging Results} Table~\ref{tab:instrinsic_eval_full} includes the full results of the quality analysis of the generated CoTs using both multi-task learning (MTL) and model merging as the combination methods for atomic CoT models. Given the same combination method, combined Composable CoT models generate responses including both atomic CoT patterns more frequently than combined atomic CoT models.

\begin{table*}
\footnotesize
\centering
\renewcommand{\arraystretch}{1.}
\begin{tabular}{llrrrr}
\toprule
 & Method & \multicolumn{1}{l}{Performance} & \multicolumn{1}{l}{\% $\mathcal{T}_{1}$ CoT} & \multicolumn{1}{l}{\% $\mathcal{T}_{2}$ CoT} & \multicolumn{1}{l}{\% Both CoT} \\
\midrule
& StandardCoT-Merge & 70.4 & 85.3 & 95.1 & 85.3 \\
Next Letter & ComposableCoT-Merge & 95.4 & 100.0 & 100.0 & $^\dagger$\textbf{100.0} \\
 + Mult  & StandardCoT-MTL & 3.6 & 0.0 & 100.0 & 0.0 \\
 & ComposableCoT-MTL & 96.3 & 98.9 & 100.0 & $^\dagger$98.9 \\
 \midrule
 & StandardCoT-Merge & 77.0 & 90.3 & 98.7 & 90.0 \\
Concat & ComposableCoT-Merge & 75.4 & 91.6 & 100.0 & \textbf{91.6} \\
+ Mult & StandardCoT-MTL & 72.1 & 99.7 & 32.1 & 32.1 \\
 & ComposableCoT-MTL & 74.3 & 100.0 & 83.1 & $^\dagger$81.3 \\
 \midrule
 & StandardCoT-Merge & 54.8 & 100.0 & 99.4 & \textbf{99.4} \\
Concat & ComposableCoT-Merge & 19.2 & 44.6 & 60.5 & 17.7 \\
 + Next Letter& StandardCoT-MTL & 60.9 & 100.0 & 66.7 & 66.7 \\
 & ComposableCoT-MTL & 63.3 & 100.0 & 85.9 & $^\dagger$85.0 \\
 \midrule
  Skill-Mix & StandardCoT-Merge & 29.8 & 60.0 & 59.2 & 35.9 \\
Literary & ComposableCoT-Merge & 39.6 & 64.1 & 66.9 & $^\dagger$\textbf{43.3} \\
+ Rhetorical & StandardCoT-MTL & 42.0 & 65.3 & 58.0 & 37.6 \\
 & ComposableCoT-MTL & 49.0 & 64.5 & 65.7 & $^\dagger$42.0 \\
 \bottomrule
\end{tabular}
\caption{Intrinsic evaluation of the generated CoTs from atomic CoT models evaluated on the compositional task in the zero-shot setting. ``\% $\mathcal{T}_{1}$ CoT'' denotes the percentage of generated responses that use the CoT format of the first atomic task of the composition, and likewise for the second. $^\dagger$ denotes that the ComposableCoT method has a significantly higher  ``\% Both CoT'' than the StandardCoT counterpart at the $0.01$ level using a paired bootstrap test. Combined Composable CoT models generate responses including both atomic CoT patterns more frequently than combined atomic CoT models.}
\label{tab:instrinsic_eval_full}
\end{table*}

\section{Details of Skill Composition for Math Reasoning}
\label{appx:details_math}
\subsection{Setup}
We follow \citet{he2025skilltargetedadaptivetraining} to evaluate ComposableCoT for math reasoning. Models are fine-tuned on a training set of math word problems and their solutions in CoT format, each of which is designed to demonstrate the use of a specific skill for mathematical reasoning. Then, they are evaluated on the test set of the MATH dataset to solve math word problems. \citet{he2025skilltargetedadaptivetraining} show that models fine-tuned with such skill-targeted data perform better on the MATH test set and other OOD math reasoning datasets than models fine-tuned on the vanilla MATH training set and other SFT datasets that are not designed to teach math reasoning skills. We use the training dataset \emph{STAT-Syn} created by \citet{he2025skilltargetedadaptivetraining} and formulate the setting as a skill composition problem.

\paragraph{Atomic tasks} Following the definitions in \citet{metacognitive}, \citet{he2025skilltargetedadaptivetraining} curate a list of mathematical skills for each math subject that are useful for problem solving. We treat these as atomic reasoning skills for the purpose of this study. Each training example in \textit{STAT-Syn} is designed to feature a single skill, making it ideally suited for our setting.

For each math subject, we select the five most frequently used skills in the training dataset to filter out skills that do not have enough training examples. The 7 subjects and the 5 selected skills for each are the following.
\begin{itemize}
    \item \textbf{Algebra:} (1) Algebraic manipulation; (2) calculation and conversion; (3) arithmetic; (4) solving equations; (5) algebraic expression.
    \item \textbf{Counting and Probability:} (1) Number theory and arithmetic operations; (2) combinatorial mathematics; (3) counting principles;
    (4) probability concepts and calculations; (5) understanding and applying combinatorics concepts.
    \item \textbf{Geometry:} (1) Algebraic; (2) triangle geometry; (3) 3D geometry and volume calculation; (4) understanding circle properties and algebraic manipulation; (5) coordinate geometry and transformation.
    \item \textbf{Intermediate Algebra:} (1) Polynomial; (2) understanding and application of functions; (3) simplification and basic operations; (4) algebraic manipulation and equations; (5) complex number manipulation and operations.
    \item \textbf{Number Theory:} (1) Basic arithmetic; (2) understanding of fractions; (3) number manipulation; (4) number theory; (5) modular arithmetic.
    \item \textbf{Prealgebra:} (1) Counting and number theory; (2) solving linear equations; (3) basic arithmetic operations; (4) multiplication and division; (5) fractions and decimals.
    
    \item \textbf{Precalculus:} (1) Trigonometric calculations; (2) geometric relations; (3) algebra and equations; (4) vector operations; (5) geometry and space calculation.
\end{itemize}

An example of the atomic CoT data can be found in Example~\ref{example:atom:algebra}.

\begin{prompt}[title={\thetcbcounter{} Atomic Task Example: STAT-Algebra Example}, label=example:atom:algebra]
\char`[Subject\char`]\\
Algebra
\\
\char`[Instruction\char`]\\
Problem: Find $h(x)$, with terms in order of decreasing degree, if $3x^4+2x-1+h(x)=5x^2-6x-1$.\\Answer:\\

\char`[Chain-of-Thought + Answer String\char`]\\
This equation is solved by $h(x)=(5x^2-6x-1)-(3x^4+2x-1)$=\\boxed\char`{$-3x^4+5x^2-8x$\char`} \\

\char`[Skill\char`]\\
Solving Equation

\end{prompt}

\paragraph{Compositional tasks} We consider each test example in the MATH dataset as a compositional example of the math reasoning skills discussed above. We follow \citet{he2025skilltargetedadaptivetraining} to annotate the required skills for each MATH test example using GPT-4o-2024-11-20. An example of the compositional data can be found in Example~\ref{example:comp:algebra}.

\begin{prompt}[title={\thetcbcounter{} Compositional Task Example: STAT-Algebra Example}, label=example:comp:algebra]
\char`[Subject\char`]\\
Algebra\\

\char`[Instruction\char`]\\
Problem: What is the smallest value of $x$ such that $|5x - 1| = |3x + 2|$? Express your answer as a common fraction.\\Answer:\\

\char`[Chain-of-Thought + Answer String\char`]\\
There are two cases, when $5x-1=3x+2$ and when $5x-1=-(3x+2).$ The two equations yield $x=\frac{3}{2}$ and $x=-\frac{1}{8},$ respectively, of which $x=$boxed\char`{$-\frac{1}{8}$\char`} \\ is the smaller solution. \\

\char`[Skill\char`]\\
Algebraic manipulation,
Solving equations,
Arithmetic

\end{prompt}

\paragraph{Difference from the Controlled Settings} We note that the mathematical skill composition setting above is not a controlled compositional setting, as the string manipulation, arithmetic, and natural language skill composition settings in Section~\ref{sec:exp_setup} are, for the following reasons. 

(1) Atomic skills are not clearly separated from each other. As seen in the list of atomic skills above, many atomic skills overlap with each other within the same subject (e.g., ``algebraic manipulation'' and ``algebraic expression'' in the algebra subject). 

(2) Compositions might be seen during pre-training. The task performance cannot be completely attributed to the composition of learned atomic skills, as the pre-trained capabilities of math reasoning might have also played a role.

\begin{table}[]
\centering
\begin{tabular}{ccc}
\toprule
Subject & \# Train Examples &  \\
\midrule
Algebra & 250 &  \\
Counting and Probability & 300 &  \\
Geometry & 100 &  \\
Intermediate Algebra & 400 &  \\
Number Theory & 200 &  \\
Prealgebra & 100 &  \\
Precalculus & 200 &  \\
\bottomrule
\end{tabular}
\caption{Training data distribution of the math skill composition setting.}
\label{tab:math_data_condition}
\end{table}

\subsection{Training}
\paragraph{Standard CoT Data} We use the \textit{STAT-Syn} training dataset from \citet{he2025skilltargetedadaptivetraining} and select the training examples that demonstrate the skills in the list above. We note that the original \textit{STAT-Syn} dataset includes training examples of duplicate CoTs, and training on such examples makes models memorize specific CoT data verbatim. Thus, we pre-process the \textit{STAT-Syn} data by de-duplication. The filtered dataset includes $100 - 500$ training examples for each subject, which is consistent with our low-data setting in Section~\ref{sec:exp_setup}. The distribution can be found in Table~\ref{tab:math_data_condition}. We use the filtered dataset as the standard CoT dataset.

\paragraph{Composable CoT Data} To augment the standard CoT dataset, we use the same data construction implementation used for two-way and three-way compositions of the string manipulation, arithmetic, and natural language skill tasks; see Section~\ref{sec:exp_setup} and Appendix~\ref{appx:details_three_way}.

\paragraph{Training Models} We fine-tune an atomic CoT model \emph{for each subject}, using the same training configurations described in Appendix~\ref{appx:hypers}. We use multi-task learning to combine ComposableCoT models and StandardCoT models.

\subsection{Inference}
At inference time, we use the same inference scheme and configuration for the three-way compositions of the string manipulation, arithmetic, and natural language skill tasks, as described in Appendix~\ref{appx:details_three_way}. Recall that at inference time, we assume knowledge of how many skills are required (i.e., the number of compositions) for each test example in the MATH test set, which ranges from 1 to 6-way compositions. We only evaluate each atomic CoT model on the MATH test set of the corresponding subject that it has been trained on; we do not perform cross-subject evaluation, as the atomic skills are subject-specific.

\subsection{Evaluation}
\paragraph{Evaluation Metric} We use the lm-eval-harness library \citep{eval-harness} to run all evaluations, and we evaluate all models on the test split of MATH following \citet{he2025skilltargetedadaptivetraining}. We use the answer parser from \citet{hendrycks2021measuring, eval-harness, lighteval} to extract answers from the model-generated CoTs, and report exact match accuracy as the evaluation metric.

\paragraph{Performance by Subject} Recall that we fine-tune a single model for each subject in MATH, since the atomic skills are specific to each subject. Therefore, we evaluate each subject-specific fine-tuned model on the test set of \emph{only the corresponding subject}. We report the weighted average across subjects as the overall accuracy, because different subjects in the MATH test set have different numbers of test examples.

\paragraph{Performance Breakdown} To evaluate the performance breakdown by the number of required skills, as reported in Table~\ref{tab:math_zero_shot}, we follow \citet{he2025skilltargetedadaptivetraining} to annotate the required skills for each MATH test example by GPT-4o-2024-11-20, and investigate the performance breakdown across test examples that: (1) Use at least 2 of the learned atomic skills; (2) Require 2--5 skills. We note that $6\%$ of the test examples are not included in the performance breakdown rows, because they either use atomic skills that are not learned, or use 1 or more than 5 skills.

\section{Error Analyses}
\label{appx:error_analysis}
In addition to not being able to perform the individual atomic task correctly, we show three types of common errors made by ComposableCoT variants in the zero-shot compositional evaluation setting.
\begin{enumerate}
    \item Example~\ref{example:err:repeated_cot} shows an example where the generated CoT is only able to replicate CoT of one atomic CoT and repeat the same CoT in the prefix and suffix. 
    \item Example~\ref{example:err:one_cot} shows an example where the combined model fails to continue generation after generating the prefix CoT. This is a common error for Composable models combined with model merging. 
    \item Example~\ref{example:err:wrong_order} shows an example where the combined model uses the wrong atomic CoT in the prefix that should have been used in the suffix. 
\end{enumerate}

\begin{prompt}[title={\thetcbcounter{} Error Case: Replicating One Atomic CoT Pattern}, label=example:err:repeated_cot]

\char`[Instruction\char`]\\
Take the first letter of each word in the sequence, concatenate them in lower case, and find the next letter in alphabet following the last letter in the sequence of the concatenated letters: Jessika Berk Hamilton Stephanie answer:

\char`[Answer String\char`]\\
t

\char`[Model Output\char`]\\
<prefix> The last letter is e, and the letter following it in alphabet is f. So the answer is f.</prefix> <suffix> The last letter is e, and the letter following it in alphabet is f. So the answer is f.</suffix>

\char`[Predicted Answer\char`]\\
f
\end{prompt}

\begin{prompt}[title={\thetcbcounter{} Error Case: Failing to Continue Generation after Prefix CoT}, label=example:err:one_cot]

\char`[Instruction\char`]\\
Take the last letter of each word in the sequence, concatenate them in lower case, then find the ASCII value of the last letter in the sequence of the concatenated letters, and multiply the ASCII value by 6: Ayat Amaris Heavenly Mariela answer:

\char`[Answer String\char`]\\
582

\char`[Model Output\char`]\\
<prefix> The last letter of the 1st word is t. The last letter of the 2nd word is s. The last letter of the 3rd word is y. The last letter of the 4th word is a. So the answer is tasy, and the ASCII value of the last letter in the sequence of the concatenated letters is 121, so the answer is 726.</prefix> <suffix>

\char`[Predicted Answer\char`]\\
726
\end{prompt}

\begin{prompt}[title={\thetcbcounter{} Error Case: Use the Wrong Order of Atomic CoTs}, label=example:err:wrong_order]

\char`[Instruction\char`]\\
Take the second letter of each word in the sequence, concatenate them in lower case, and find the next letter in alphabet following the last letter in the sequence of the concatenated letters: D'shaun Takia Hiroki Arjun answer:

\char`[Answer String\char`]\\
s

\char`[Model Output\char`]\\
<prefix> the last letter is n, and the letter following it in alphabet is o. so the answer is o.</prefix> <suffix> the second letter of d is d, the second letter of t is a, the second letter of h is i, and the second letter of a is r. so the answer is dair.</suffix>

\char`[Predicted Answer\char`]\\
dair
\end{prompt}

\section{Computation resources}\label{sec:compute-res}

We conducted all our experiments primarily on a server with NVIDIA RTX A6000 48GB GPU, and a server with GPU of NVIDIA GH200 120GB. Each individual experiment can be run on a single A6000 of 48 GB within 1 hour.

\section{Licensing}
\label{appx:license}
We use the following publicly available datasets from prior works with open licenses.
\paragraph{Skill-Mix} \citet{yu2024skillmix} use the Apache-2.0 license and data is available at: 
\url{https://huggingface.co/spaces/dingliyu/skillmix}.

\paragraph{Letter concatenation} The dataset uses the Apache-2.0 license and the data is available at: \url{https://huggingface.co/datasets/ChilleD/LastLetterConcat}

\paragraph{STAT-Syn} \citet{he2025skilltargetedadaptivetraining} use the Apache-2.0 license and data is available at: 
\url{https://github.com/princeton-pli/STAT}.

\end{document}